\documentclass[10pt,twocolumn,letterpaper]{article}

\usepackage[pagenumbers]{cvpr} %

\usepackage{bbold}
\usepackage{bbold}

\usepackage{graphicx}

\usepackage{xspace}
\usepackage{enumitem}

\usepackage{duckuments} %
\usepackage{subcaption}
\usepackage[dvipsnames]{xcolor}
\usepackage{diagbox} %
\usepackage{xstring}
\usepackage{multirow}
\usepackage{arydshln}

\usepackage{amssymb}%
\usepackage{pifont}%

\usepackage{caption} %
\usepackage{fp}
\usepackage{expl3}[2012-07-08]
\ExplSyntaxOn
\ExplSyntaxOff

\usepackage{amsmath,amsfonts,bm}

\def\x{{x}}

\def\xi{{\x_i}}

\newcommand{\ignorethis}[1]{}
\newcommand{\myparagraph}[1]{\vspace{-2pt} \smallskip \noindent \textbf{#1}}

\def\eqref#1{equation~\ref{#1}}

\def\1{\bm{1}}

\DeclareMathAlphabet{\mathsfit}{\encodingdefault}{\sfdefault}{m}{sl}
\SetMathAlphabet{\mathsfit}{bold}{\encodingdefault}{\sfdefault}{bx}{n}

\newcommand{\ignore}[1]{}

\makeatletter
\DeclareRobustCommand\onedot{\futurelet\@let@token\@onedot}
\def\@onedot{\ifx\@let@token.\else.\null\fi\xspace}
\def\eg{e.g\onedot,\xspace}

\def\etal{\emph{et al}\onedot}
\makeatother

\newcommand*{\menlo}{\fontfamily{lmtt}\fontsize{8}{8}\selectfont }

\definecolor{MyDarkBlue}{rgb}{0,0.08,1}
\definecolor{MyDarkGreen}{RGB}{27,89,0}
\definecolor{MyDarkRed}{rgb}{0.8,0.02,0.02}
\definecolor{MyDarkOrange}{rgb}{0.40,0.2,0.02}
\definecolor{MyPurple}{RGB}{111,0,255}
\definecolor{MyRed}{rgb}{1.0,0.0,0.0}
\definecolor{MyGold}{rgb}{0.75,0.6,0.12}
\definecolor{MyDarkgray}{rgb}{0.66, 0.66, 0.66}
\definecolor{myorange}{RGB}{255,69,0}

\newcommand{\supp}[1]{#1}%
\newcommand{\revision}[1]{\color{black}#1\color{black}}

\definecolor{cvprblue}{rgb}{0.21,0.49,0.74}
\usepackage[pagebackref,breaklinks,colorlinks,allcolors=cvprblue]{hyperref}

\def\paperID{3828} %
\def\confName{CVPR}
\def\confYear{2025}

\title{Identifying Prompted Artist Names from Generated Images}

\author{
Grace Su$^{1}$ \hspace{10mm}
Sheng-Yu Wang$^{1}$ \hspace{10mm}
Aaron Hertzmann$^{2}$ \\
\hspace{-8mm}
Eli Shechtman$^{2}$ \hspace{9mm}
Jun-Yan Zhu$^{1}$ \hspace{16mm}
Richard Zhang$^{2}$ \\
\\
$^{1}$Carnegie Mellon University \hspace{10mm} 
$^{2}$Adobe Research
}

\begin{document}

\twocolumn[{%
\renewcommand\twocolumn[1][]{#1}%
\maketitle
\vspace{-5mm}
\begin{center}
    \centering
    \includegraphics[width=1.\linewidth]{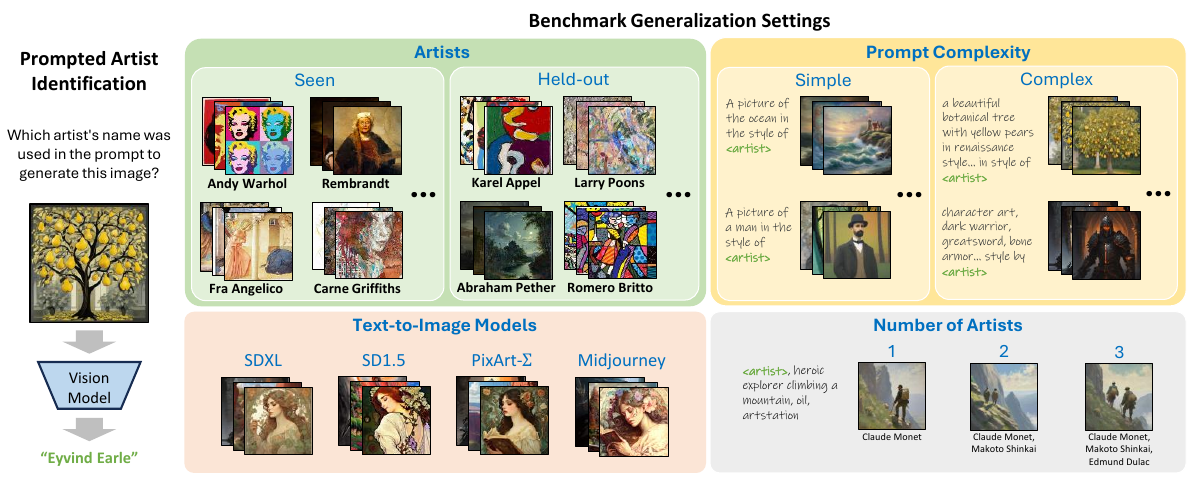}
    \captionof{figure}{
    \textbf{Prompted Artist Identification Benchmark}.  
    We introduce the first large-scale benchmark for identifying prompted artist names from generated images. The benchmark covers four axes of generalization that match realistic use cases: 
   \textbf{(1) Artists:} we collect artists commonly used in prompts and simulate open-set artist classification by testing on artists not seen during training. \textbf{(2) Prompt complexity:} users describe images in many different ways, including short, simple prompts as well as more descriptive, complex prompts. \textbf{(3) Text-to-image models:} users can generate images using various text-to-image models, which have different training data and architectures that may affect the generated image's overall style. \textbf{(4) Number of artists:} users may include multiple artists in the prompt to mix styles, creating images that are not easily attributable to a single artist. 
    }
\label{fig:teaser}
\end{center}
}]

\begin{abstract}
A common and controversial use of text-to-image models is to generate pictures by explicitly naming artists, such as ``in the style of Greg Rutkowski''. 
We introduce a benchmark for \emph{prompted-artist recognition}: predicting which artist names were invoked in the prompt from the image alone.  
The dataset contains 1.95M images covering 110 artists and spans four generalization settings: held-out artists, increasing prompt complexity, multiple-artist prompts, and different text-to-image models. 
We evaluate feature similarity baselines, contrastive style descriptors, data attribution methods, supervised classifiers, and few-shot prototypical networks. Generalization patterns vary: supervised and few-shot models excel on seen artists and complex prompts, whereas style descriptors transfer better when the artist’s style is pronounced; multi-artist prompts remain the most challenging.  
Our benchmark reveals substantial headroom and provides a public testbed to advance the responsible  moderation of text-to-image models. We release the dataset and benchmark to foster further research: \href{https://graceduansu.github.io/IdentifyingPromptedArtists}{https://graceduansu.github.io/IdentifyingPromptedArtists}.
\end{abstract}

\begin{figure*}[t]
    \centering
    \includegraphics[width=1.\linewidth]{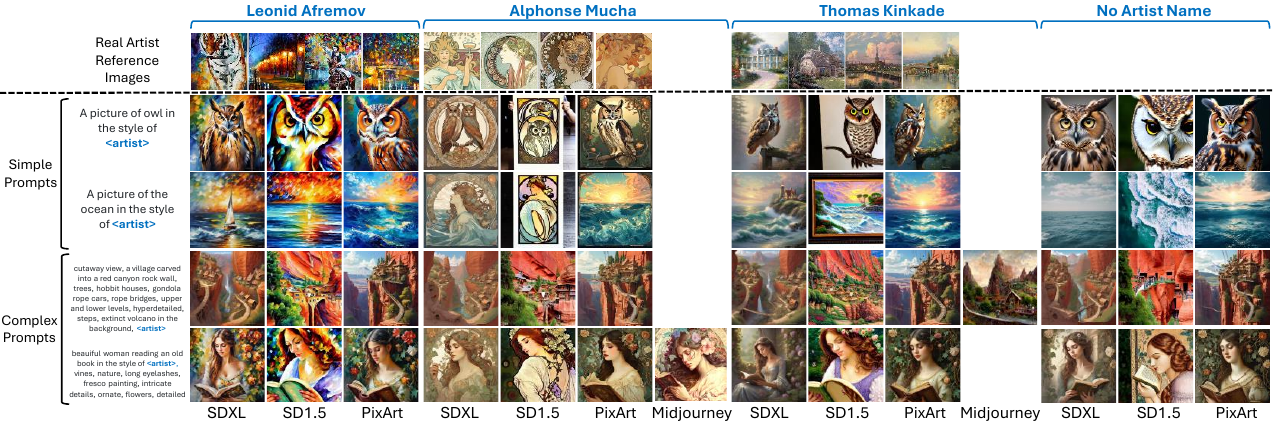}
    \caption{\textbf{Prompted Artist Identification Dataset}. 
    We construct a structured dataset of 1.95M images to benchmark different methods on predicting prompted artist names from generated images. 
    To help disentangle the effect of prompting given an artist name, we query a text-to-image model given the same content prompt (rows), but insert different artist names (columns). Our dataset consists of images generated by SDXL~\cite{podell2023sdxl}, SD1.5~\cite{rombach2022high}, PixArt-$\Sigma$~\cite{chen2024pixart}, and Midjourney~\cite{midjourney},
    with both complex and simple prompts. 
    Each artist's style tends to become less visually prominent when the prompt becomes more complex, especially when the prompt calls for additional styles and adjectives or specifies the content that may not be in the distribution of content in the artist's work (e.g., 2nd row, ``a village carved into a red canyon rock wall''). The variance in the visibility of the artist's style across different prompts and models demonstrates the difficulty of the prompted artist identification task.}
\label{fig:dataset}
\end{figure*}

\section{Introduction}
\label{sec:intro}
Current text-to-image models allow users to generate images specifying artistic styles, often by direct reference to artists' names, such as `in the style of a named artist'. 
Some online artwork sharing platforms prohibit uploading such imagery~\cite{adobe2023,GettyImages2024}, due to potential harm to artists~\cite{andersen2022,rutkowski2022,matsakis2023} and the risk of producing derivative works that appear very similar to,  or indistinguishable from, the original artist's works.
However, without access to the original prompt, it is unclear how to detect such violations.

In this work, we focus on the problem of automatically classifying which artist's name was directly used in a prompt. 
We refer to this problem as identifying the ``prompted artist'' from the generated image. To tackle this problem, we introduce a large-scale benchmark of 1.95 million labeled images. 
Our benchmark is designed to evaluate generalization across four axes that reflect real-world use of text-to-image models for generating images derived from existing artists' styles, as illustrated in Figure~\ref{fig:teaser}.

\revision{
\textbf{(1) Artists.}
The set of artists that may be included in a prompt is open-ended. Any given model is likely to encounter images generated with artists that were not included in its training set, so we create a set of held-out artists in the benchmark.}

\revision{
\textbf{(2) Prompt complexity:} Users construct prompts in a variety of ways, from short, simple prompts to long, complex prompts that may specify additional styles, reducing the visibility of the artist's style in the resulting image. Thus, we manually design simple prompts (e.g., ``picture of the ocean in the style of...''), along with long, complex prompts scraped from databases of real use-cases~\cite{sun2024journeydb}.
We then hold out a set of test prompts from the training set to evaluate generalization across unseen prompts and content. To reduce the association of a given artist name with specific content, we query the model using the same content prompt, with different artist names inserted into the prompt.}

\revision{
\textbf{(3) Text-to-image models:} Users can choose from a large selection of popular image generators. The complex interaction of the network architecture, learning algorithm, training images, and input prompts leads to varying representations of the same artist's style that may be challenging to identify. %
We collect training and evaluation set images from commonly-used models: SDXL~\cite{podell2023sdxl}, SD1.5~\cite{rombach2022high}, PixArt-$\Sigma$~\cite{chen2024pixart}, and Midjourney~\cite{midjourney}.}

\revision{
\textbf{(4) Number of artists per prompt:} Users may include multiple artists in one prompt, creating combined styles that can be challenging to associate with any one artist. To study these cases, we create subsets for images generated with 2 and 3 artists in our benchmark. 
}

\revision{We evaluate a wide range of methods on our benchmark: feature similarity methods, contrastive style descriptors, data attribution methods, supervised classifiers, and few-shot prototypical networks.
The degree of generalization varies across different methods.
Our benchmark reveals substantial room for improvement across all generalization settings evaluated. Supervised and few-shot models trained on images generated with prompted artists perform better on seen artists and complex prompts. However, style descriptors trained on real artwork generalize better on simple prompts, where the artist's style is more apparent. We find that capturing and recognizing the representation of prompted artists, as learned and expressed by a generative model, is a related, yet distinct problem from style recognition of real artwork.
Meanwhile, prompts referencing multiple artists continue to pose the greatest challenge.
We release the benchmark and dataset to help advance the responsible moderation of AI-generated content at \href{https://graceduansu.github.io/IdentifyingPromptedArtists}{https://graceduansu.github.io/IdentifyingPromptedArtists}.
}

\section{Related Work}
\label{sec:related_work}
\myparagraph{Generated Image Detection.} 
While generated images have become increasingly photorealistic, they continue to leave detectable traces, back with manipulated faces~\cite{roessler2019faceforensicspp,zi2020wilddeepfake,dolhansky2020deepfake,li2020celeb,jiang2020deeperforensics,khalid2021fakeavceleb}, GANs \cite{marra2018detection,marra2019gans,nataraj2019detecting,zhang2019detecting,yu2019attributing,Wang_2020_cnndet,chai2020makes}, and now with diffusion models \cite{corvi2023detection,ojha2023towards,sha2023defake,wang2023dire,epstein2023online,zhu2023genimage,cazenavette2024fake,cozzolino2024raising,park2024community,hong2024wildfake}. Despite these advances, robustness to in-the-wild perturbations and generalization to future generators are a continuous challenge~\cite{Wang_2020_cnndet,cazenavette2024fake}.  Here, we study a distinct aspect: we seek not only to detect if something is generated, but also to study properties that led to its generation, i.e., if an artist's style was targeted for replication. 

\myparagraph{Data Attribution.} 
While most generated images are ``novel'' and not an exact copy of a training image, all reflect some specific elements of the training set. Data attribution seeks to link a synthesized image to the constituent training images that influenced it~\cite{wang2023evaluating,georgiev2023journey}, assuming the set of training images and training process is known, i.e., a closed world. However, for our artist name classification task, we may not know the training set and learning algorithm. Moreover, existing data attribution algorithms are computationally expensive~\cite{koh2017understanding,grosse2023studying,georgiev2023journey,zheng2024intriguing,choe2024data,wang2024attributebyunlearning,ko2024mirrored,isonuma2024unlearning}. In contrast, our benchmark focuses on finding cases where an artist name's influence is direct and intentional. We show that style descriptors and prototype classifiers outperform data attribution methods on prompted artist identification.

\myparagraph{Style Similarity.} 
Artistic style classification methods are typically trained on real artwork and extract or learn image features that are similar across images of the same style~\cite{karayev2013recognizing, saleh2015large, mao2017deepart, strezoski2018omniart, sabatelli2018deep, ruta2021aladin}.
Some recent works additionally measure the style similarity between a generated image and the artist's original work to determine whether the generated image replicates the artist's style~\cite{somepalli2024measuringstyle, moayeri2024rethinking, liu2024lora, casper2023measuring, kumar2024introstyle}. 
However, \revision{when images are generated from complex prompts that obscure the artist’s style, style similarity methods become less effective than a classifier trained directly on our dataset.}

\myparagraph{Generated Image–Text Datasets.}
\revision{
Many large-scale datasets containing paired prompts and generated images have been released,
including DiffusionDB~\cite{wang2023diffusiondb}, JourneyDB~\cite{sun2024journeydb}, and TWIGMA~\cite{chen2023twigma}, text-to-image model evaluation benchmarks~\cite{saharia2022photorealistic, yuscalingparti, lee2023holistic, bakr2023hrs, huang2025t2i}, and user preference alignment datasets~\cite{xu2023imagereward, kirstain2023pick, chen2024tailored, vodrahalli2024artwhisperer}.
Other efforts focus on user-generated art from text-to-image models and provide richer stylistic variety~\cite{zheng2024stylebreeder,wu2023goya,silva2024artbrain}. 
Yet none of these datasets targets images explicitly prompted with artist names. The closest work, by Leotta~\etal~\cite{leotta2023notwithmyname}, represents an early attempt to tackle the challenging problem of inferring artist names from generated images. They offer a dataset of 8,519 DALL$\cdot$E 2~\cite{ramesh2022hierarchical} images covering five artists. In contrast, we construct a large-scale benchmark of 1.95M images, which spans hundreds of artists, multiple generative models, diverse prompts, and varying numbers of artists, enabling comprehensive evaluation of open-set artist-name recognition.}

\begin{figure*}[!h]
    \centering
    \includegraphics[width=1.\linewidth]{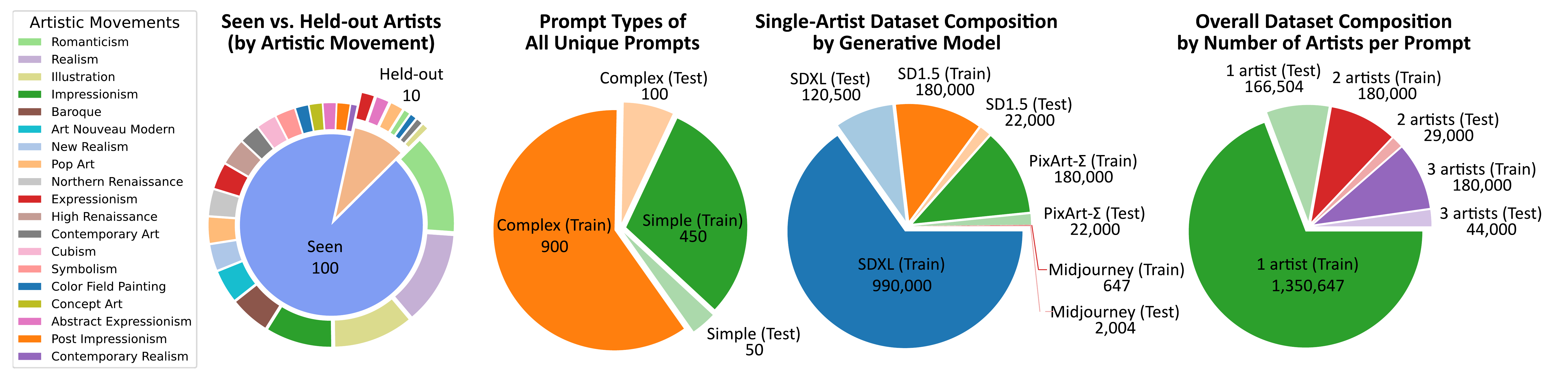}
    \caption{\textbf{Dataset Statistics}. 
    The prompted artist identification benchmark employs a structured dataset of 1.95M images to evaluate vision methods across four axes of generalization.
    We collect 110 of the most frequently prompted artists, split into 100 seen and 10 held-out artists ($1^{\text{st}}$ chart).
    Next, we collect 1,000 complex prompts and 500 simple prompts in which artist names are inserted ($2^{\text{nd}}$ chart). For seen artists, we use a separate set of prompts for testing. 
    For held-out artists, we further divide the test prompts to generate a set of reference images used during inference.
    Then, we generate single artist-prompted images with SDXL~\cite{podell2023sdxl}, SD1.5~\cite{rombach2022high}, and PixArt-$\Sigma$~\cite{chen2024pixart}, and collect Midjourney images~\cite{midjourney} ($3^{\text{rd}}$ chart).
    Finally, we evaluate how well methods generalize to multiple artists in the prompt by generating datasets of SDXL images prompted with 2 artists and 3 artists ($4^{\text{th}}$ chart).
    }
\label{fig:dataset_statistics}
\end{figure*}

\section{Prompted Artist Identification Benchmark}
\label{sec:dataset}
\revision{
Our goal is to create a benchmark evaluating the generalization of prompted artist identification methods across four relevant axes to cover the various images generated from typical prompts from real-world text-to-image model users.
Specifically, the benchmark's dataset includes seen vs. held-out artists (Section~\ref{sec:dset_artists}), prompt types of varying complexity (Section~\ref{sec:dset_prompt}), different text-to-image models (Section~\ref{sec:dset_unseen_models}), and different numbers of artists per prompt (Section~\ref{sec:dset_multi_artist}).
We refer to images generated with prompts invoking one or more artist names as ``artist-prompted images.''
}

\subsection{Seen and Held-Out Artists}
\label{sec:dset_artists}
\revision{
In the real world, users can reference an open-ended and continuously growing set of artists in image generation prompts, increasing the likelihood that a prompted artist identification model will encounter names not seen during training. While the training set may cover the most frequently used artists, the ability to generalize to held-out artists without retraining remains important. 
To ensure the benchmark includes the most relevant artists for real-world applications, we collect a list of artists most commonly used in text-to-image model prompts, then designate a set of held-out artists to evaluate how well vision models generalize to unseen artists. 
}

Specifically, we manually de-duplicate and filter 110 artists from an initial list of 400 artists frequently prompted by Stable Diffusion users on the \url{Lexica.art} website~\cite{somepalli2024measuringstyle, lexicaartstablediffusionprompts}. 
For these artists, we then obtain 10k real reference images from LAION-Styles~\cite{somepalli2024measuringstyle}, a curated subset of LAION-5B~\cite{schuhmann2022laion} targeted towards artist style attribution. 
To evaluate how well various methods generalize to different artists, we use 100 artist names for training (``seen'') and 10 distinct artists for testing (``held-out'')
Our real artist image dataset thus includes 9907 images from seen artists and 860 images from held-out artists.
\supp{We include the artist list, curation details, and artist name frequencies in our LAION subset in the supplement.}

\subsection{Varying Prompt Complexity}
\label{sec:dset_prompt}
Text-to-image models have the flexibility to generate images with text prompts of varying lengths and complexities. Intuitively, for shorter prompts, the artist's name influences the style of the resulting images more heavily, while longer, more complex prompts may dilute this effect. Hence, we study how classification performance differs in the following two prompt types.

\myparagraph{Simple prompts.}
We generate images with simple prompts in the form of ``{\menlo A picture of <content> in the style of <artist>}.''  For diversity, we use 500 different contents sampled from ChatGPT~\cite{chatgpt}. Figure~\ref{fig:dataset} shows examples of the generated images; they typically follow a consistent artist style. 

\myparagraph{Complex prompts.}
In a realistic setting, a user often generates images with more complex prompts, such as ``{\menlo Pisces zodiac sign as a cat alphonse mucha art style realistic 3d render attention to details hyper realistic}.'' The above prompt is taken from JourneyDB~\cite{sun2024journeydb}, a dataset containing text prompts that real users submitted to Midjourney~\cite{midjourney}. We collect 1000 text prompts from JourneyDB, each mentioning one artist's name. Then, we replace the artist's name with one of the 110 artists in our dataset list. For example, the previous text prompt in our dataset would be ``{\menlo Pisces zodiac sign as a cat <artist> art style realistic 3d render attention to details hyper realistic}.'' In Figure~\ref{fig:dataset}, we observe that the artist's style becomes less prominent in the generated image compared to the simple prompts, making classification a much more challenging problem.

\myparagraph{Generalize to unseen prompts.}
To evaluate how different artist classification methods generalize to different prompts, we use 450 simple prompts and 900 complex prompts for training, and hold out 50 simple prompts and 100 complex prompts for testing. \supp{We include the list of prompts and processing procedure in the supplement.}

\subsection{Different Text-to-Image Models}
\label{sec:dset_unseen_models}
\myparagraph{Curating generators and images.} 
Many text-to-image generative models have been developed in recent years with varying degrees of open-source access~\cite{jiang2024genaiarena, lee2023holistic, li2025k, zhao2024evolvedirector, liesenfeld2024rethinking}, and more continue to be released as the field progresses.
Generated images may come from any of these models, each of which renders prompts and artist names differently, depending on the model's training data and architecture.
Thus, our benchmark also evaluates how well prompted artist identification methods generalize to different text-to-image models.
To curate a large, structured image dataset labeled with the original generation prompts, we select recent, popular models that are 1) able to generate artistic styles when an artist name is directly prompted for and 2) have open-source weights or a large published dataset of image-prompt pairs.

We tried generating images prompted with artist names using the recent SD3.5~\cite{esser2024scaling} and Flux~\cite{flux2024} open-source models, but found that using artist names often has little to no effect on the generated image style, even when the prompt is simple.
This may be because the models' training datasets were filtered to exclude problematic content~\cite{stabilityai_safety}. While the models may be able to replicate the target style if the user writes prompts describing the style, they do not respond to artist names as well as earlier models like SDXL~\cite{podell2023sdxl}.
\supp{We show examples of their generated images in the supplement.} 

Thus, for each artist and prompt combination, we generate images with 2 seeds using the open-weight models SDXL~\cite{podell2023sdxl}, SD1.5~\cite{rombach2022high}, and PixArt-$\Sigma$~\cite{chen2024pixart}. For SDXL images generated with complex prompts, we use 10 different generation seeds. 
To collect images from Midjourney~\cite{midjourney}, a closed-source model, we filter the JourneyDB dataset~\cite{sun2024journeydb} for our set of seen and held-out artists. The resulting dataset is summarized by Figure~\ref{fig:dataset_statistics}.
\supp{Full dataset statistics tables are included in the supplement.}

\subsection{Multiple Artists per Prompt}
\label{sec:dset_multi_artist}
\begin{figure}[h]
    \centering
    \includegraphics[width=\linewidth]{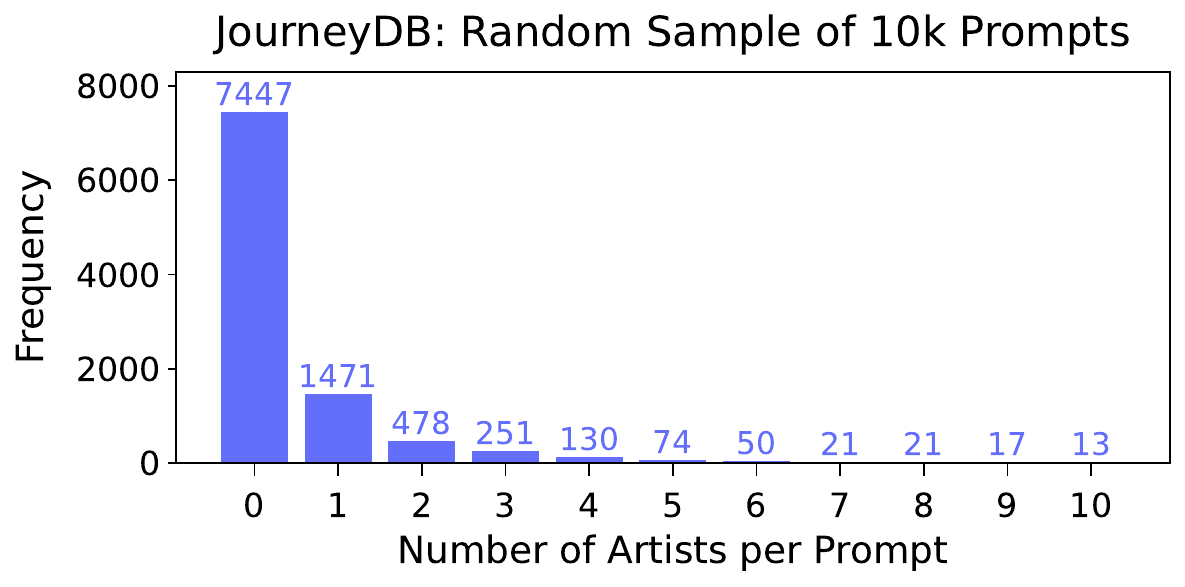}
    \vspace{-5pt}
    \caption{\textbf{Statistics on number of artists per prompt.} 
    To understand how frequently text-to-image model users prompt for multiple artists, we take a random sample of 10,000 prompts collected from Midjourney users in JourneyDB~\cite{sun2024journeydb} and visualize the distribution of the number of artists per prompt.
    While 14.7\% of the prompts invoke only one artist name, 4.8\% contain 2 artist names, and 2.5\% contain 3 artist names.
    }
    \label{fig:num_artists_per_prompt}
\end{figure}

Text-to-image model users not only prompt for a single artist's style, but also mix multiple artists' styles by including multiple artist names in the prompt. For instance, in a random sample of 10,000 prompts collected from Midjourney users in JourneyDB~\cite{sun2024journeydb}, we find that 10.8\% of the prompts contain multiple artist names, compared to 14.7\% containing 1 artist name and 74.5\% containing no artist name. The distribution, visualized in Figure~\ref{fig:num_artists_per_prompt}, 
is long-tailed and heavily skewed, with 4.8\% of the prompts containing 2 artist names, 2.5\% containing 3 artist names, and 1.3\% or less for 4 or more artist names.

Thus, our benchmark primarily contains single-artist prompted images for evaluation. However, we also evaluate how well prompted artist identification methods generalize to multiple artists by generating a dataset of images prompted with 2 artists, and a dataset of images prompted with 3 artists. 

\myparagraph{Generating images prompted with multiple artists.}
To curate a dataset of multi-artist prompted images that allow for evaluation of generalization to held-out artists and varying prompt complexity, we slightly modify the procedure described in Section~\ref{sec:dset_prompt}. We insert multiple artist names into the prompt templates, \eg ``{\menlo Pisces zodiac sign as a cat <artist 1> and <artist 2> art style realistic 3d render attention to details hyper realistic}'', sample 100 random seen artist combinations and all possible held-out artist combinations for 2 artists and 3 artists, and generate images with SDXL. 
The multi-artist datasets are summarized by the fourth chart in Figure~\ref{fig:dataset_statistics}. 

\subsection{Image Similarity Analysis}
\label{sec:dset_similarity}
In Figure~\ref{fig:dataset}, we qualitatively observe that the generated images become less aligned with the style of the artist's original work when the prompt is more complex. In addition, given the same prompt, the style of the generated image varies depending on which text-to-image model was used.
Thus, we conduct a simple quantitative analysis of CLIP~\cite{radford2021learning} image similarity scores and find three aspects that make the prompted artist identification task challenging for generalization:
1) The effect of the artist's name in the prompt on the generated image is diluted when the prompt is more complex.
2) The alignment of the generated images to the artist's average artwork varies across different text-to-image models.
3) The effect of adding artists into the prompt diminishes as the number of artists increases.

\myparagraph{Comparing style alignment of text-to-image models.} 
Different text-to-image models have different training datasets and architectures, which may affect the generated image's overall style.
Therefore, we compare the text-to-image models in our benchmark by comparing image similarities across different text-to-image models in Table~\ref{tab:clip_similarity}.
For each artist-prompted image, we compute the CLIP image similarity between it and the image generated with the same seed and prompt, except that the artist's name is removed.
For all models, we observe that the CLIP similarity scores are lower for simple prompts than for complex prompts, indicating that the artist name has a larger influence on the generated image when the prompt is simple. 
Prompting for an artist's name also has a smaller effect on the generated image when using PixArt, compared to SDXL and SD1.5.

We also compute the similarity between the artist-prompted image embedding and the average embedding of the real artist images, which is the artist's prototype.
When the prompt is more complex, the generated images are less aligned with the artist's real style.
In addition, PixArt images are less aligned to the artist's real style than SDXL and SD1.5. 

\begin{figure*}[t]
    \centering
    \includegraphics[width=\linewidth]{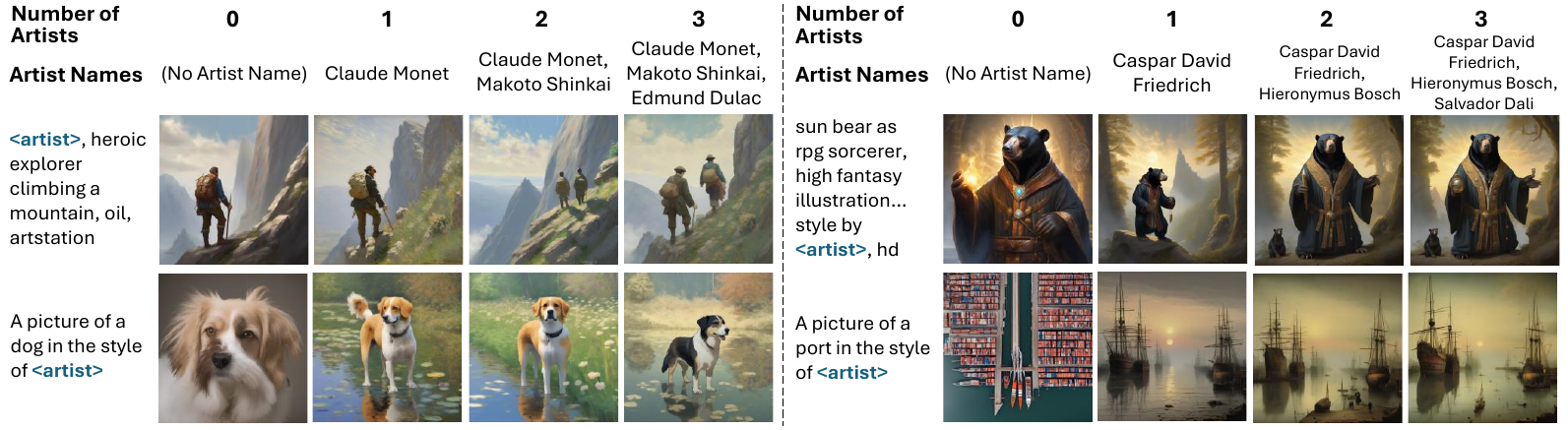}
    \caption{\textbf{Multi-artist prompted images}. 
    In the example images shown, we observe that the effect of adding each artist's name in the prompt diminishes as more artists are added. Each row shows a set of images generated with the same prompt and generation seed, and the number of artists inserted into the prompt increases from left to right. For each prompt, the image changes less with each additional artist, and images generated with complex prompts generally change less than those generated with simple prompts. 
    }
\label{fig:multi_artist_dataset}
\end{figure*}

\myparagraph{Influence of adding artist names into prompts.}
Since image generation prompts may contain more than one artist name, we study the effect of increasing the number of artist names in the prompt and how difficult it is to identify all the artists.
In Table~\ref{tab:similarity_num_artists}, we show that the similarity between each image generated with the same seed and prompt increases as we add more artist names into the prompt, for both complex and simple prompts. 
We can qualitatively observe this decrease in influence per added artist in Figure~\ref{fig:multi_artist_dataset}. 
This indicates that classifying multiple artists in a prompt is more difficult than classifying a single artist, because the effect of each artist name is diluted as more artist names are added to the prompt.

\begin{table*}[!t]
\footnotesize
\centering
\hfill
\begin{subtable}[t]{0.4\textwidth}
    \centering
    \caption{\textbf{Text-to-image model comparison}}
    \resizebox{\linewidth}{!}{
    \begin{tabular}{lcccc}
    \toprule
    \multirow{3}{*}{Model} & \multicolumn{2}{c}{\shortstack[c]{Content only vs.\\Artist-prompted}} & \multicolumn{2}{c}{\shortstack[c]{Artist-prompted vs.\\Real prototype}} \\
    \cmidrule(lr){2-3} \cmidrule(lr){4-5}
     & Simple & Complex & Simple & Complex \\
    \midrule
    SDXL       & 0.571 & 0.738 & 0.570 & 0.476 \\
    PixArt    & 0.632 & 0.816 & 0.522 & 0.441 \\
    SD1.5      & 0.520 & 0.643 & 0.590 & 0.527 \\
    Midjourney & --   & --   & --   & 0.541 \\
    \bottomrule
    \end{tabular}
    }
    \label{tab:clip_similarity}
\end{subtable}
\hspace{4mm}
\begin{subtable}[t]{0.35\textwidth}
    \centering
    \caption{\textbf{Increasing number of artists}}
    \resizebox{\linewidth}{!}{
    \begin{tabular}{lccc}
    \toprule
    & \multicolumn{3}{c}{Number of Artists Compared} \\
    \cmidrule(lr){2-4}
    Prompt Type & 0 vs. 1 & 1 vs. 2 & 2 vs. 3 \\
    \midrule
    Complex & 0.738 & 0.811 & 0.866 \\
    Simple  & 0.571 & 0.711 & 0.782 \\
    \bottomrule
    \end{tabular}
    }
    \label{tab:similarity_num_artists}
\end{subtable}
\hfill
\vspace{-2mm}
\caption{\textbf{Average CLIP image similarities}. (a) For each text-to-image model, we compute the average CLIP similarity between the image generated without an artist name and the image generated with one artist name (first two columns), and the average CLIP similarity between the image generated with an artist name and the real prototype for that artist (last two columns). 
The artist's influence and generated image's alignment to the artist's real style decreases when the prompt is more complex. Additionally, PixArt images are less aligned to the artist's real style than SDXL and SD1.5 images. 
(b) Given the same prompt and generation seed, we compute the average CLIP similarity between SDXL images generated with 0, 1, 2, and 3 artist names. The influence of each artist's name in the prompt diminishes as more artist names are added. 
}
\label{tab:combined_clip_similarity}
\end{table*}

\section{Experimental Setup}
\label{sec:exp_setup}
We evaluate a variety of computer vision models with published implementations on our prompted artist identification benchmark and employ evaluation procedures that use the same train and test sets across all models for fair comparison.

\subsection{Models}
\label{sec:models}
Two approaches to classifying generated images are searching for similar artist reference images and training feed-forward classifiers. While retrieving image features from models trained on large-scale datasets allows for generalization to unseen classes, they require more runtime and storage compared to feed-forward classifiers. Thus, we compare several pre-trained retrieval-based methods and classifiers trained on our prompted artist identification dataset. \supp{We include full training details and ablations in the supplement.}

\begin{itemize}[leftmargin=12pt]
  \setlength{\parskip}{0pt}
  \setlength{\itemsep}{1pt}
    \item \textbf{Contrastive Style Descriptors (CSD)~\cite{somepalli2024measuringstyle}} is a style descriptor that measures style similarity between images. CSD is trained using supervised contrastive learning~\cite{khosla2020supervised} on $\sim\hspace{-2mm}500$k curated real artist images from LAION-5B~\cite{schuhmann2022laion}. 
    \item \textbf{DINOv2~\cite{oquab2023dinov2}, CLIP~\cite{radford2021learning}:} commonly-used self-supervised image features trained on internet images. 
    \item \textbf{Attribution by Customization (AbC)~\cite{wang2023evaluating}:} DINO (AbC) and CLIP (AbC) are features for attributing training data in customized text-to-image diffusion models, fine-tuned from DINO~\cite{caron2021emerging} and CLIP~\cite{radford2021learning}, respectively. They are trained on images synthesized from SD1.5 models customized on ImageNet~\cite{deng2009imagenet} and art datasets.
    \item \textbf{Prototypical Network:} 
    We train a classifier based on Prototypical Networks~\cite{snell2017prototypical}, a few-shot learning method, which allows for the prediction of unseen artists. During training, it learns an image encoder that outputs features in the embedding space close to the correct artist's prototype, the averaged pre-trained CLIP features of the artist's real reference images. During inference, the same prototype features can be used to predict the artist label set seen during training. We can also predict artists not seen during training by using the held-out artists' reference images for prototype features.
    \item \textbf{Vanilla Classifier:} We add an MLP with one hidden layer as the classification head to the CLIP image encoder, then train all layers on our dataset. 
\end{itemize}

\subsection{Evaluation Framework}
\label{sec:eval_framework}
\myparagraph{Single-artist classification.}
We evaluate each model's top-1 classification accuracy for seen and held-out artists. For every test set of our prompted artist identification benchmark, each model is given the same set of training/retrieval images for controlled comparison. The training image set is all generated seen artist training images from our SDXL, SD1.5, PixArt, and Midjourney datasets and includes both simple and complex-prompted images. 
To evaluate retrieval-based methods on held-out artists, we use images generated with held-out prompts for test-time reference.
For the vanilla classifier, we only report seen artist classification as the model cannot be applied to classes not seen during training. To estimate the statistical significance of each evaluation, we bootstrap each model's predictions by resampling the evaluation images, with replacement, by artist name, prompt template, and generation seed for 2000 iterations. \supp{We plot our bootstrapping procedure's convergence at 2000 iterations in the supplemental material.}

\myparagraph{Multi-artist classification.}
To evaluate artist classification on images prompted with multiple artist names, we use a multi-label classification metric: ranked mean average precision for the top 10 unique retrieved labels (mAP@10). We adapt the prototypical network's training to a multi-label objective by using binary cross-entropy loss with label smoothing. The training dataset includes all single-artist images from SDXL, as well as all 2-artist and 3-artist training images. To evaluate retrieval-based methods, we only use single-artist SDXL images as the reference database. 

\begin{figure*}[!ht]
    \centering
    \includegraphics[width=\linewidth]{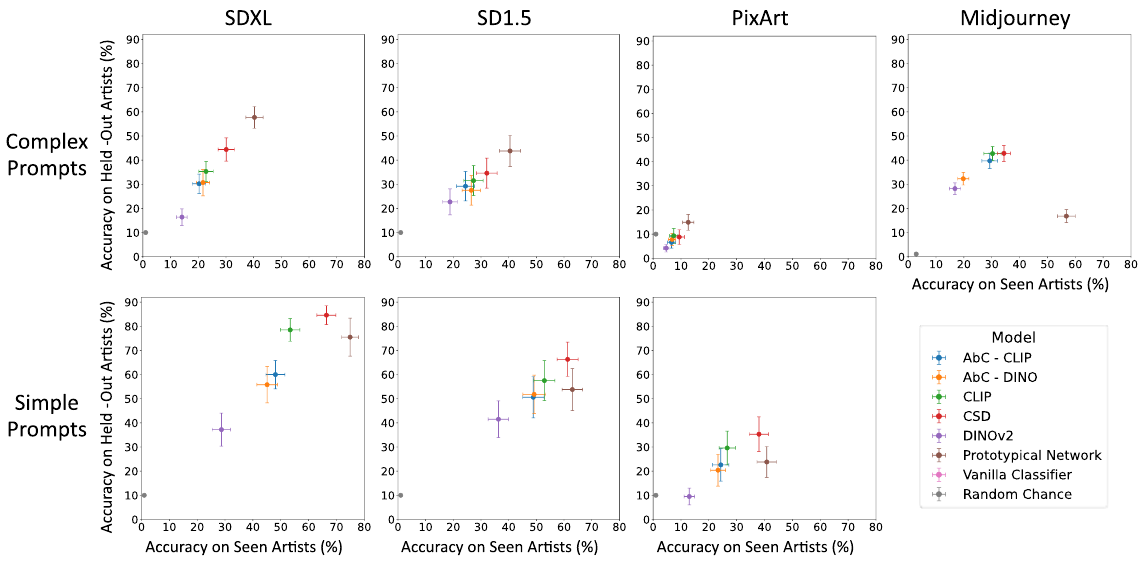}
    \caption{\textbf{Single-artist prediction results}. 
    We compare the prompted artist classification accuracy of various visual representation methods.
    We test within our seen artists set (100-way classification, x-axis) and on held-out artists (10-way classification, y-axis). We also test on images generated with different text-to-image models (SDXL, SD1.5, PixArt, and Midjourney), and complex and simple prompts. 
    Although prototypical networks~\cite{snell2017prototypical}, the trained vanilla classifier, and CSD~\cite{somepalli2023diffusion} surpass the CLIP~\cite{radford2021learning}, DINOv2~\cite{caron2021emerging}, and AbC~\cite{wang2023evaluating} methods, attaining high accuracy across all scenarios remains difficult.
    \supp{We include all numbers as tables in the supplemental material.}
    }
\label{fig:accuracy_bar_plots_single_artist}
\vspace{-2mm}
\end{figure*}

\begin{figure*}[!ht]
    \centering
    \includegraphics[width=\linewidth, trim=0 0 0 1.5cm, clip]{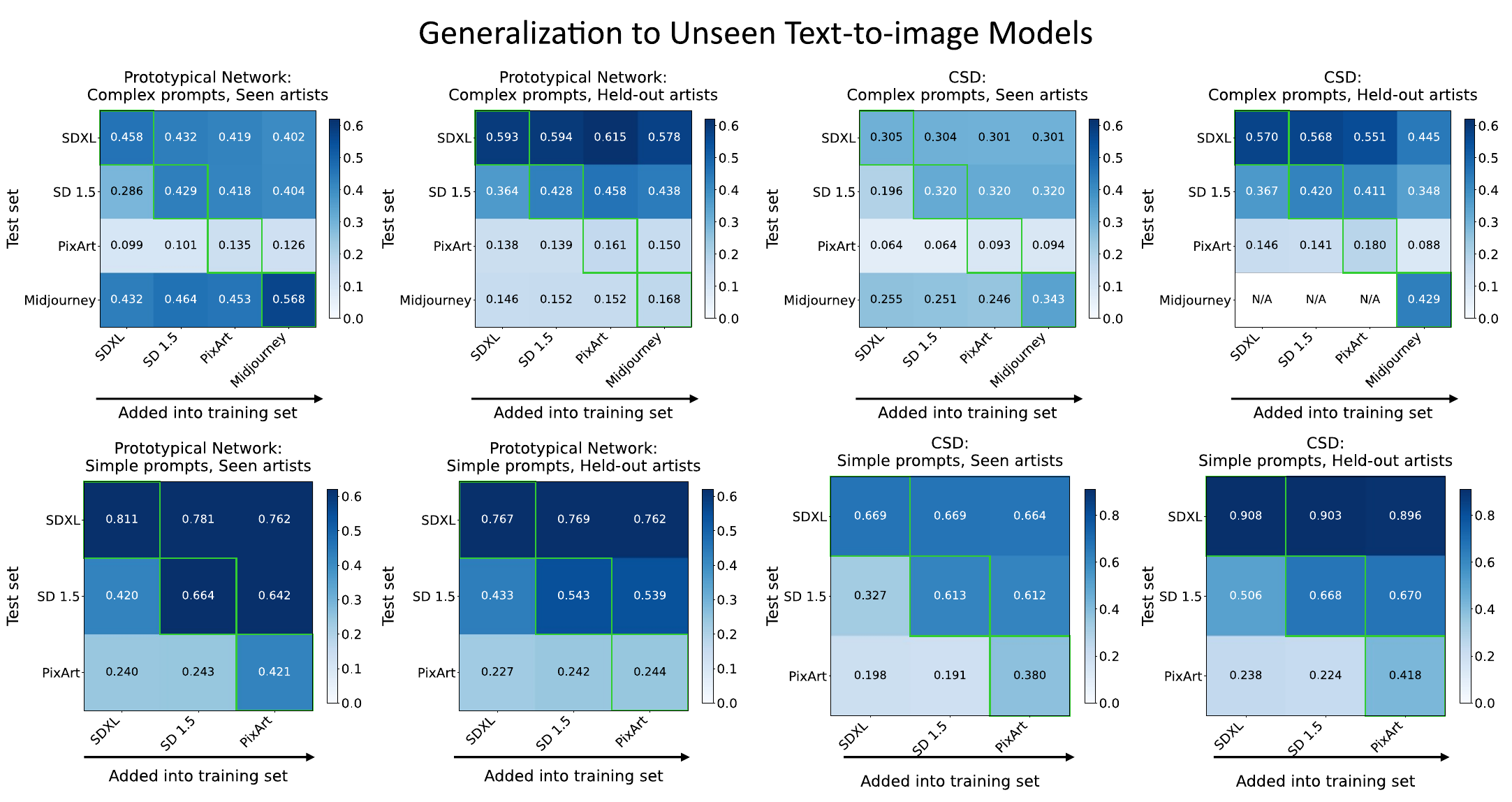}
    \caption{\textbf{Generalization to unseen generators}. 
    For the best-performing methods, prototypical networks and CSD, we evaluate performance on unseen text-to-image models by progressively increasing the training set of images. 
    Cells below the green diagonals evaluate generalization to unseen domains. 
    For both methods, expanding the training dataset does not enhance performance on unseen text-to-image models; improvements occur only when the training data includes images generated by the specific model being evaluated.
    }
\label{fig:online_learning_generator_prototype}
\vspace{-2mm}
\end{figure*}

\begin{figure}[!ht]
    \centering
    \includegraphics[width=\linewidth]{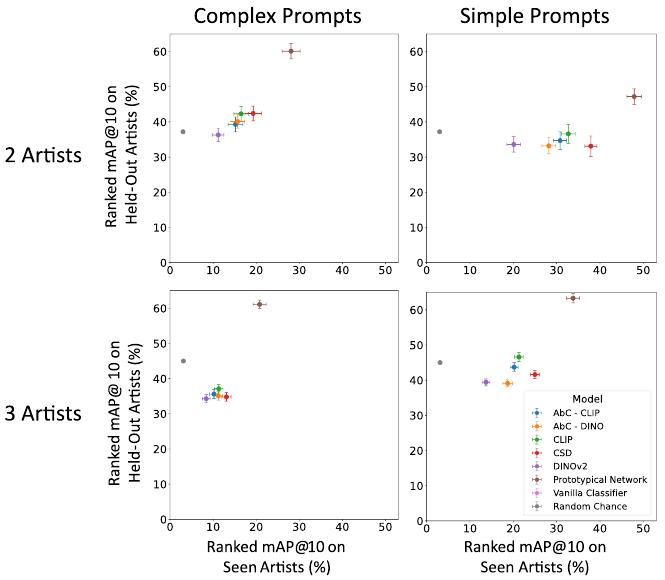}
    \caption{\textbf{Multi-artist prediction results}. We evaluate visual representation methods on the multi-artist classification task, where the input image is prompted with multiple artists' names. We test on SDXL-generated images with 2 artists and 3 artists in the prompt, and report ranked mAP@10 on seen artists (x-axis) and held-out artists (y-axis).
    All methods generally exhibit reduced performance on images generated from complex prompts compared to those generated from simple prompts.
    As expected, the prototypical network achieves the highest performance across the datasets due to its training on multi-artist prompted images. However, its performance falls short of saturation.
    \supp{We include all numbers as tables in the supplemental material.}
    }
\label{fig:accuracy_bar_plots_multi_artist}
\vspace{-2mm}
\end{figure}

\begin{figure}[!ht]
    \centering
    \includegraphics[width=\linewidth, trim=0 0 0 1.5cm, clip]{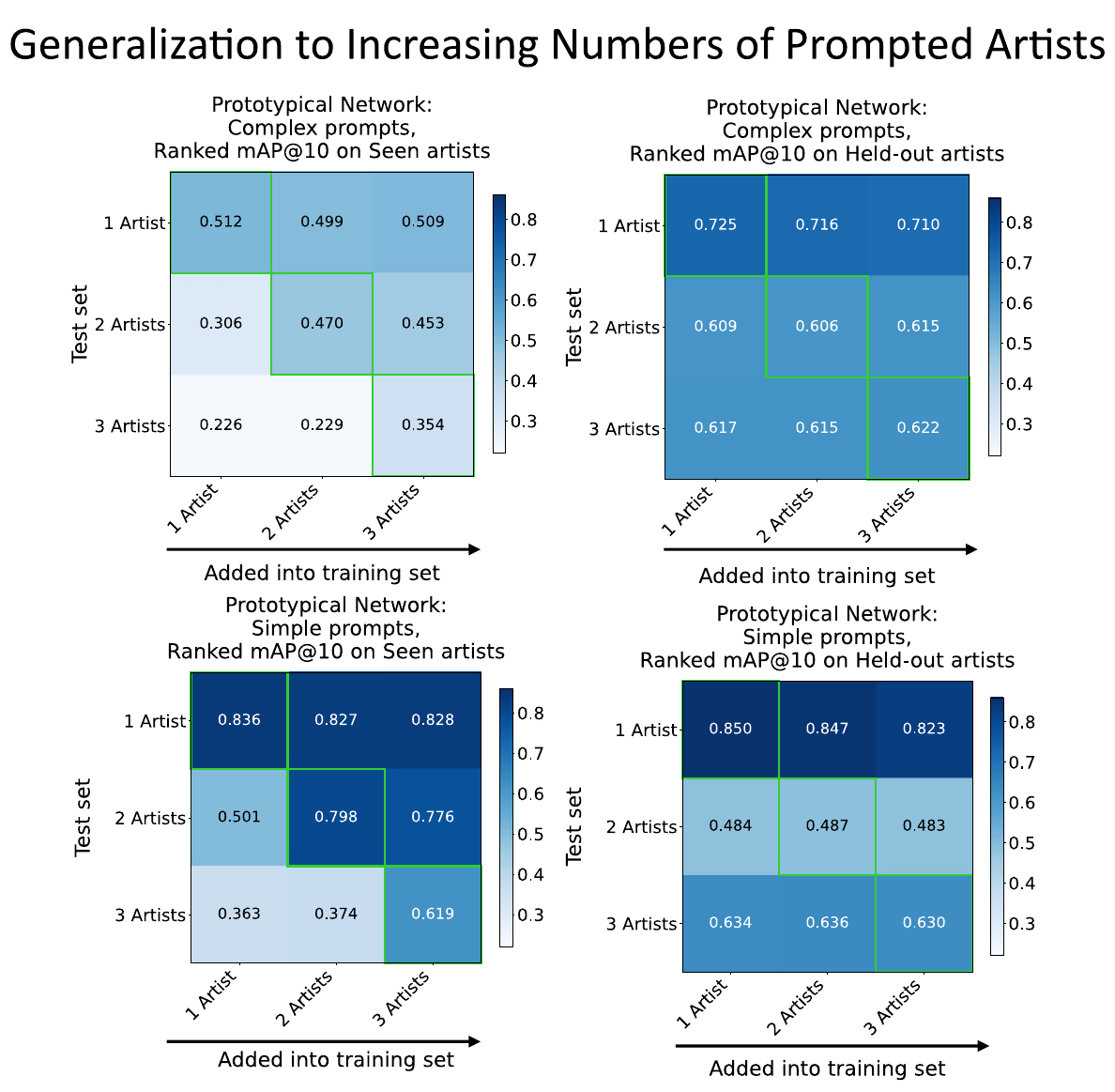}
    \caption{\textbf{Generalization to increasing numbers of prompted artists}. 
    For prototypical networks, we evaluate performance on unseen numbers of artists in the prompt by progressively increasing the training set of images to include multi-artist prompted images.
    Cells below the green diagonals evaluate generalization to unseen domains. 
    Increasing the size of the training set does not improve generalization to unseen numbers of artists. While using training images with the same number of artists as the test set enhances classification for seen artists, it does not benefit performance on held-out artists.
    }
\label{fig:online_learning_multiartist}
\vspace{-2mm}
\end{figure}

\section{Results}
\label{sec:results}
We analyze prompted artist classification performance of various visual representation methods on all generalization settings of our benchmark.

\subsection{Single-artist Classification}
We present the quantitative results of all methods on the prompted artist classification task in Figure~\ref{fig:accuracy_bar_plots_single_artist}. While all methods surpass random chance, none exceed 91\% accuracy, underscoring the inherent difficulty of the task. Performance consistently declines with increasing prompt complexity and when evaluating on PixArt-generated images, compared to SDXL and SD1.5-generated images. This suggests that \textit{the complexity of the prompt and the text-to-image model used for image generation substantially influence classification difficulty.}

Across all generalization tasks, CSD, prototypical networks, and vanilla classifiers are the best-performing methods, followed by CLIP, the AbC models, and finally DINOv2. This indicates that \textit{style descriptors and trained classifiers are more effective than off-the-shelf general-purpose visual representations and data attribution methods} for prompted artist identification. 

Prototypical networks outperform CSD on most evaluation datasets and can learn a representation of the artist's style from generated images that is more robust to complex prompts than CSD, which is trained on real artworks. 
Meanwhile, CSD generalizes better to images with held-out artists and simple prompts, and Midjourney images with complex prompts and held-out artists. 
This indicates that \textit{complex prompts generate images that are less aligned to the artist's real style compared to simple prompts.}

\myparagraph{Generalization to unseen text-to-image models.}
We evaluate the generalization of the best-performing methods, prototypical networks, and CSD, to unseen text-to-image models by progressively increasing the training set of images~\cite{epstein2023online} to include images from additional generators (SDXL$\rightarrow$SD1.5$\rightarrow$PixArt$\rightarrow$Midjourney).
Results are shown in Figure~\ref{fig:online_learning_generator_prototype}. For both methods, increasing the training dataset does not improve performance on unseen text-to-image models---performance only improves on each text-to-image model when the training set includes images from that generator. This suggests that \textit{the models have not learned style representations that generalize across text-to-image models}. Notably, performance on PixArt images does not improve much, even after including PixArt images in the training set, supporting our observation in Section~\ref{sec:dset_similarity} that PixArt images are less aligned to the artist's real style compared to SDXL or SD1.5. 

\subsection{Multi-artist Classification}
We further evaluate all methods on the multi-artist classification task, where the input image is prompted with multiple artists' names, and show results in Figure~\ref{fig:accuracy_bar_plots_multi_artist}. 
\textit{Across methods, performance tends to decline on images produced using complex prompts compared to those produced with simple prompts.}
The prototypical network is the best performing method on the multi-artist datasets, as expected, because it is the only method trained on multi-artist prompted images, but performance remains far from saturation. The vanilla classifier is the second-best performing, even though it is not trained on multi-artist prompted images, followed by CSD. The remaining retrieval-based methods perform similarly to CSD.

\myparagraph{Generalization to increasing numbers of prompted artists.}
We also examine the impact of adding multi-artist prompted images to the training set of the prototypical network in Figure~\ref{fig:online_learning_multiartist}. Expanding the training set does not lead to better generalization to unseen artist counts. Training with images containing the same number of artists as the test set improves seen artist classification but not held-out artist classification. This indicates that \textit{the prototypical network is not able to learn a representation of the artist's style that generalizes to unseen numbers of artists and held-out artists}, even when trained on multi-artist prompted images. The image similarity analysis in Section~\ref{sec:dset_similarity} shows that the images generated with multiple artists are more similar to each other than images generated with a single artist, which may be the reason for this lack of generalization.

\begin{figure}[h]
    \centering
    \includegraphics[width=\linewidth]{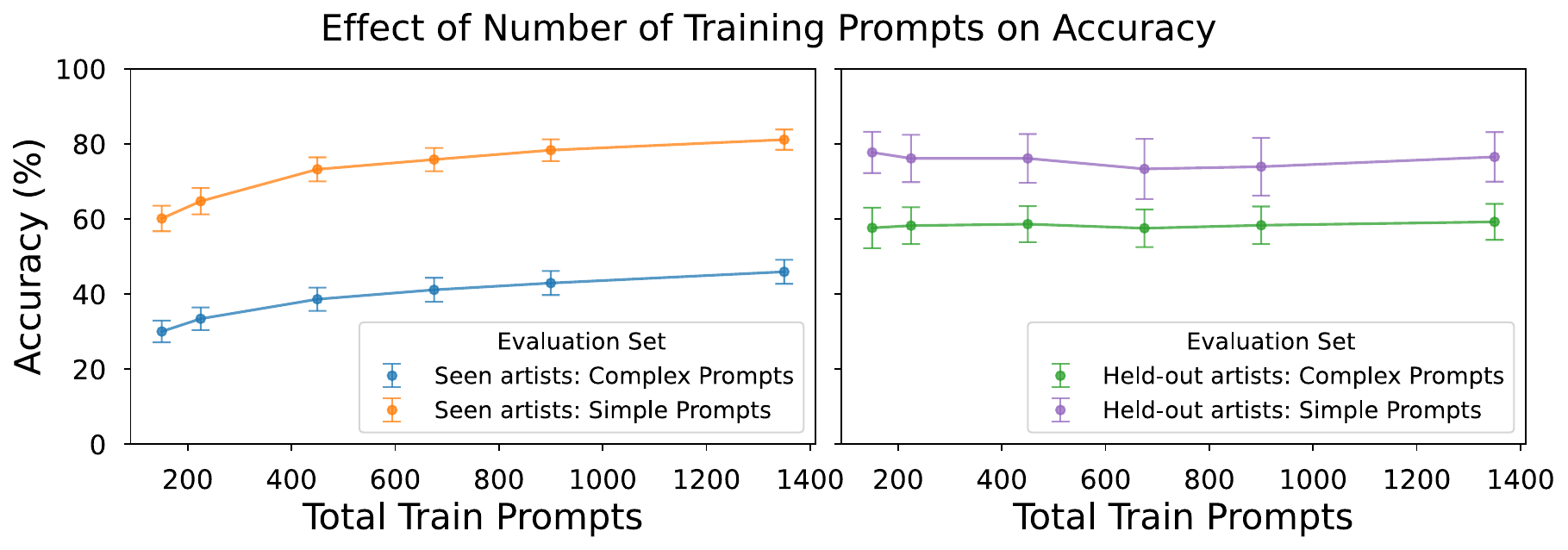}
    \caption{\textbf{Ablation on number of training prompts}. We plot the prototypical network classification accuracy on seen artists (left) and held-out artists (right) as a function of the number of prompts in the training set. Classification accuracy on seen artists improves steadily with more training prompts, but performance on held-out artists remains unchanged across all dataset sizes. The prototypical network is able to learn a representation of the artist's style that generalizes to unseen prompts with seen artists, but not to prompts with held-out artists. 
    }
\label{fig:ablation_train_prompts}
\vspace{-2mm}
\end{figure}

\subsection{Additional Results}
\label{sec:additional_analysis}
\myparagraph{Training dataset ablation.}
To assess the effect of training dataset size, we vary the number of training prompts for the prototypical network from 150 to 1350 while keeping the ratio of simple to complex prompts fixed at 1:2. As shown in Figure~\ref{fig:ablation_train_prompts}, classification accuracy on seen artists consistently improves with more training prompts. In contrast, performance on held-out artists remains largely unchanged. This indicates that \textit{while training the prototypical network on more unique prompts improves the model's representations of styles from seen artists, it does not improve generalization to unseen artists.}

\myparagraph{Artist name detection.}
Predicting whether an image was prompted with an artist name at all is another related task that is useful for generated content moderation applications.
Thus, we evaluate several top-performing methods on this binary classification task. We construct a dataset from our artist-prompted SDXL subset, augment it with prompts that do not contain artist names, and balance the test set as shown in Table~\ref{tab:dataset_detection}.

We evaluate linear probe classifiers on the CLIP, CSD, and prototypical network representations, and a fully fine-tuned CLIP model. 
Results in Table~\ref{tab:detection_results} show that while all methods exceed random chance, none achieve perfect accuracy even though they are fine-tuned for the detection task. 
Performance is consistently higher on simple prompts, aligning with our observation in Table~\ref{sec:dset_similarity} that simple prompts yield greater feature differences between content-only and artist-prompted images---making them easier to distinguish than complex prompts.
Notably, the fully fine-tuned CLIP model under-performs on the test set, indicating susceptibility to overfitting. 
This is likely because \textit{the task is inherently harder with more intra-class variability}, as the model must learn to distinguish between images that are prompted with an artist's name and images that are not, and within each class, there is low visual similarity.

\myparagraph{Detecting non-public domain artists.}
An important downstream application of prompted artist identification is detecting improper usage, such as the generation of derivative works based on copyright-protected artists. 
We evaluate the performance of various methods on the task of detecting non-public domain artists' names by grouping the SDXL-prompted artist identification dataset into two meta-labels: public domain or non-public domain, as defined by 95 years post-mortem. We test on the 100-way seen artists test set, which contains 55 public domain and 45 non-public domain artists, and report performance in Table~\ref{tbl:precision_non_public_domain}. 
All methods perform better than random chance, but the trained classifier methods---prototypical networks and vanilla classifiers---perform the best.
This indicates that \textit{given a closed set of artists, training methods on artist-prompted images can improve the detection of non-public domain artists.}

\begin{table}[!ht]
\centering
\resizebox{\linewidth}{!}{%
\begin{tabular}{cccccc}
\toprule
\multicolumn{1}{l}{} & \multicolumn{1}{l}{} & \multicolumn{2}{c}{Train}             & \multicolumn{2}{c}{Test}              \\ \cmidrule(lr){3-4} \cmidrule(lr){5-6}
\multicolumn{1}{l}{} & \multicolumn{1}{l}{} & Content only & Artist-prompted & Content only & Artist-prompted \\ \hline
\multirow{4}{*}{\shortstack[c]{Complex\\Prompts}} & \# Seen artists     & --     & 100     & --     & 5     \\
                                 & \# Held-out artists & --     & 0       & --     & 5     \\
                                 & \# Prompts          & 900   & 900     & 100   & 100   \\
                                 & \# Seeds            & 10    & 10      & 10    & 1     \\ \cdashline{1-6}
\multirow{4}{*}{\shortstack[c]{Simple\\Prompts}}  & \# Seen artists     & --     & 100     & --     & 5     \\
                                 & \# Held-out artists & --     & 0       & --     & 5     \\
                                 & \# Prompts          & 450   & 450     & 50    & 10    \\
                                 & \# Seeds            & 2     & 2       & 2     & 1     \\ \hline
\multicolumn{1}{l}{}             & \# Images           & 9,900 & 990,000 & 1,100 & 1,100 \\ \bottomrule
\end{tabular}%
}
\caption{\textbf{Artist name detection dataset}. To evaluate different methods on the task of detecting whether an image was prompted with an artist name, we construct a dataset of SDXL images prompted with artist names and images prompted with the same content but without the artist names. The training set contains all artist-prompted SDXL images from our benchmark, and the test set is balanced between content-only and artist-prompted images. }
\label{tab:dataset_detection}
\end{table}

\begin{table}[!ht]
\centering
\resizebox{\linewidth}{!}{%
\begin{tabular}{lcccccc}
\toprule
 \multirow{2}{*}{Method} 

                                    & \multicolumn{3}{c}{Complex}                                     & \multicolumn{3}{c}{Simple}                                      \\ \cmidrule(lr){2-4} \cmidrule(lr){5-7}
 &
  \multicolumn{1}{c}{Accuracy} &
  \multicolumn{1}{c}{AP} &
  \multicolumn{1}{c}{AUCROC} &
  \multicolumn{1}{c}{Accuracy} &
  \multicolumn{1}{c}{AP} &
  \multicolumn{1}{c}{AUCROC} \\ \hline
Chance                              & 50.0          & -- & -- & 50.0          & -- & -- \\ \hline
CLIP - linear probe                 & 61.6          & 79.9                & 79.5                & 82.5          & 97.7                & 97.9                \\
CSD - linear probe                  & 70.2          & 81.3                & 79.6                & \textbf{92.5} & \textbf{98.2}       & \textbf{98.3}       \\
Proto. Net. - linear probe & \textbf{70.7} & \textbf{83.6}       & \textbf{80.8}       & 91.0          & 97.3                & 97.0                \\ 
\cdashline{1-7}
Full CLIP fine-tune & 67.3          & 78.6                & 79.3                & 84.5          & 96.4                & 97.1                \\\bottomrule
\end{tabular}%
}
\caption{\textbf{Artist name detection results}. We evaluate the best prompted artist classification methods on the artist name detection task and report accuracy, average precision (AP), and area under the receiver operating characteristic curve (AUCROC). Even though all methods are fine-tuned for the detection task, none achieve perfect accuracy. Performance is consistently higher on simple prompts, and the fully fine-tuned CLIP model under-performs on the test set, indicating susceptibility to overfitting. This is likely because the artist name detection task is inherently harder.
}
\label{tab:detection_results}
\end{table}

\begin{table}[!ht]
\centering
\resizebox{\linewidth}{!}{
\begin{tabular}{lcccccc} \toprule
\multirow{2}{*}{Method} 
           & \multicolumn{3}{c}{Complex}                              & \multicolumn{3}{c}{Simple}                               \\ \cmidrule(lr){2-4} \cmidrule(lr){5-7}
     & Accuracy                   & Precision    & Recall       & Accuracy                   & Precision    & Recall       \\ \hline
Chance     & 45.0 & --            & --            & 45.0 & --            & --           \\ \hline
CLIP       & 64.3 ± 1.6               & 63.7 ± 5.8 & 69.4 ± 2.8 & 81.0 ± 2.0               & 79.7 ± 4.2 & 84.0 ± 2.8 \\
CSD        & 68.3 ± 1.7               & 67.0 ± 5.6 & 74.1 ± 2.7 & 86.5 ± 1.8               & 85.3 ± 3.4 & 88.9 ± 2.4 \\
Proto. Net. & 74.3 ± 2.0          & 70.3 ± 5.4          & \textbf{85.2 ± 2.3} & \textbf{91.2 ± 1.6} & \textbf{90.3 ± 2.8} & 92.8 ± 2.4          \\
Classifier   & \textbf{74.4 ± 1.8} & \textbf{71.9 ± 5.3} & 81.2 ± 2.6          & 91.1 ± 1.5          & 89.9 ± 2.8          & \textbf{92.9 ± 2.1} \\ \bottomrule
\end{tabular}
}
\caption{\textbf{Accuracy on non-public domain, seen artists}. 
We evaluate the performance of various methods on the task of detecting the presence of a non-public domain artist name in the prompt. We test on the 100-way seen artists test set, which contains 55 public domain and 45 non-public domain artists. The trained classifier methods, prototypical networks and vanilla classifiers, perform the best, suggesting that training on generated images can improve detection of non-public domain artists, given a closed set of artists. 
\vspace{-3mm}
}
\label{tbl:precision_non_public_domain}
\end{table}

\section{Discussion and Limitations}
In this work, we explored the task of identifying prompted artist names in a generated image and demonstrated that it poses significant generalization challenges for current visual representation methods. We built the first large-scale benchmark and dataset suited for this task, comprised of 1.95 million images and prompted with 110 artist names. 
The benchmark is designed to probe four dimensions of generalization: held-out artists, varying prompt complexity, different text-to-image models, and prompts referencing multiple artists. 
Using this benchmark, we evaluated a diverse set of vision models. 
In summary, our findings are:
\begin{itemize}
    \item Supervised classifiers and few-shot prototypical networks~\cite{snell2017prototypical} trained on artist-prompted images
    excel on seen artists and complex prompts.
    \item Contrastive style descriptors~\cite{somepalli2024measuringstyle} transfer better to images where the artist's style is more visually apparent.
    \item Learning style recognition from original artworks does not fully capture how style is represented in generative models.
    \item Prompts invoking multiple artist names remain the most challenging. %
    \item There remains substantial room for improvement in generalization across all settings.
\end{itemize}
These findings highlight the need for vision models with improved generalization ability on prompted artist identification. 
We release our benchmark and dataset to facilitate the development of generalizable vision methods for the responsible moderation of AI-generated content.

\myparagraph{Limitations.}
We focused our benchmark on four text-to-image models as a starting point, but we acknowledge that there are several other popular closed-source models that could be evaluated and our Midjourney~\cite{midjourney} evaluation was limited in size.
We also did not evaluate on images generated with personalization techniques~\cite{ruiz2023dreambooth, kumari2022customdiffusion, ye2023ip}, another common way text-to-image users replicate artist styles. 
Finally, we did not evaluate the performance of multi-modal large language models, as many lack domain-specific knowledge about artists~\cite{bin2024gallerygpt}.

\myparagraph{Acknowledgments.}
We thank Maxwell Jones and Kangle Deng for their helpful discussion and comments. 
This work started when Grace Su was an Adobe intern. Grace Su is supported by the NSF Graduate Research Fellowship (Grant No. DGE2140739). This project was partially supported by NSF IIS-2403303, Adobe Research, and the Packard Fellowship. 

{
    \small
    \bibliographystyle{ieeenat_fullname}
    \bibliography{main}
}

\clearpage
\twocolumn[
        \centering
        \Large
        \textbf{Supplementary Material} \\
        \vspace{1.0em}
]
We include additional dataset curation details (Section~\ref{sec:dataset_details}), 
training details for the classifier methods (Section~\ref{sec:training_details}), 
relevant method experiments and ablations (Section~\ref{sec:method_exps_ablations}),
and additional evaluation details (Section~\ref{sec:eval_details}). All referenced {\menlo .txt} files are available in the released dataset at \href{https://huggingface.co/datasets/cmu-gil/PromptedArtistIdentificationDataset/tree/main/metadata}{https://huggingface.co/datasets/cmu-gil/PromptedArtistIdentificationDataset/tree/main/metadata}.

\section{Dataset Curation Details}
\label{sec:dataset_details}

\subsection{Artist Curation}
\label{sec:artist_curation}
As mentioned in Section~\ref{sec:dataset} of the main text, to collect a list of artists commonly used in text-to-image models, we manually de-duplicate and filter our set of seen artists from an initial list of 400 artists frequently prompted by Stable Diffusion users on the \url{Lexica.art} website~\cite{somepalli2024measuringstyle, lexicaartstablediffusionprompts}. To ensure our set of 10 held-out artists are unseen by CSD~\cite{somepalli2024measuringstyle} during its training, we filter LAION-5B~\cite{schuhmann2022laion} for artists from the leaked list of 16,000 artists~\cite{belci2024leakedmj} used to train Midjourney~\cite{midjourney}, that have at least four images having an aesthetic score greater than 4. 

We then cross-reference our filtered artists with the captions in CSD's dataset to ensure our held-out artists' names do not appear in them. 
We include our final list of 100 seen artists in {\menlo seen\_artists.txt}, list of 10 held-out artists in {\menlo held\_out\_artists.txt}, and list of 55 public domain artists in {\menlo public\_domain\_artists.txt}. 

\myparagraph{Artist images.}
We collect real reference images for each artist from a filtered version of LAION-5B~\cite{schuhmann2022laion} that only includes safe and legal images~\cite{laion2023safety}. Then, we follow the same curation procedure as LAION-Styles~\cite{somepalli2024measuringstyle}, filtering for images with predicted aesthetics scores of 6 or higher, image captions containing the final dataset's 3840 style tags, and images with SSCD~\cite{pizzi2022self} similarity less than 0.8. 

\myparagraph{Multi-artist images.}
To generate images with prompts containing 2 or 3 artist names, we insert all possible held-out artist combinations and 100 random seen artist combinations into each prompt. We include the 100 random combinations of 2 seen artists in {\menlo seen\_artist\_2combos\_100samples.txt}, the 100 random combinations of 3 seen artists in {\menlo seen\_artist\_3combos\_100samples.txt}, all combinations of 2 held-out artists in {\menlo held\_out\_artist\_all\_2combos.txt}, and all combinations of 3 held-out artists in {\menlo held\_out\_artist\_all\_3combos.txt}.

\subsection{Prompt Curation and Image Generation} 
\myparagraph{Simple prompts.}
We generate images with simple prompts in the form of ``{\menlo A picture of <content> in the style of <artist>}.'' 
For diversity, we use 100 subjects sampled from ChatGPT~\cite{chatgpt}. The prompt we give to ChatGPT is ``{\menlo Provide a comma-separated list of 100 subjects that are commonly featured in paintings}.'' We manually inspect and remove subjects that may be too out-of-distribution for most artists' works or abstract concepts, such as ``fairies'' or ``politics'', and then prompt ChatGPT again to generate subjects until we curate 100 subjects in total.

\myparagraph{Complex prompts.}
To curate a set of complex prompts from real text-to-image model users, we filter from the JourneyDB~\cite{sun2024journeydb} train dataset, which contains $\sim$4M prompt-image pairs. We first obtain a list of prompts that only mention one artist, rather than multiple artists, by instructing the large language model Llama 3 8B Instruct~\cite{dubey2024llama}. Specifically, for each JourneyDB prompt, we instruct Llama 3 with the following: ``{\menlo Output only the names that are artists that are mentioned in the following text as a JSON list. Do not say any extra words:}.'' Using a large language model for this task ensures we obtain more prompts that we would find with hard string matching to a closed set of artists. Then, we remove prompts where Llama 3 returns more than one artist name. We also remove prompts that are less than 5 words or have more than 77 CLIP~\cite{radford2021learning} text tokens to ensure SDXL~\cite{podell2023sdxl} is able to condition images on the full prompt. Finally, we sample 1,000 prompts for our dataset and remove near-duplicate prompts with manual inspection. 
We include our final lists of prompts with the train and test splits.
Complex prompts are in {\menlo complex\_prompts\_train.txt} and {\menlo complex\_prompts\_test.txt}, and simple prompts are in {\menlo simple\_prompts\_train.txt} and {\menlo simple\_prompts\_test.txt}. 
For evaluation on held-out artists, we use images generated with the first 5 prompts in {\menlo complex\_prompts\_test.txt} as test-time reference images for complex prompts, and image generated with the first 5 prompts in {\menlo simple\_prompts\_test.txt} as test-time reference images for simple prompts.

\myparagraph{Image generation.}
For each artist and prompt combination from our curated lists, we generate artist-prompted images using the open-source text-to-image models SDXL~\cite{podell2023sdxl}, SD1.5~\cite{rombach2022high}, and PixArt-$\Sigma$~\cite{chen2024pixart}. For each model, we use its default generation parameters: 
SD1.5 with the resolution of $512 \times 512$, guidance scale of 7.5, and 50 inference steps; PixArt-$\Sigma$ with the resolution of $1024 \times 1024$, guidance scale of 4.5, and 50 inference steps; and SDXL with the resolution of $1024 \times 1024$, guidance scale of 7.5, and 50 inference steps.

\begin{figure}[t]
    \centering
    \includegraphics[width=\linewidth, trim=0 10 0 0, clip]{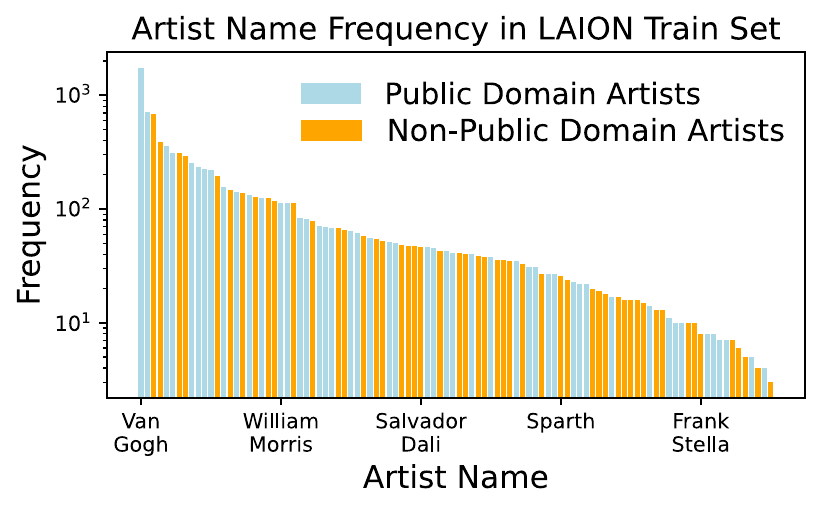}
    \vspace{-5pt}
    \caption{\textbf{Artist statistics.} We plot the frequency of each artist's name in our LAION training set, with artists labeled public domain (blue) or non-public domain (orange), as defined by 95 years post-mortem. Out of our set of 100 seen artists, 45 are non-public domain, and their images make up 38.7\% of our real image dataset.
    }
    \label{fig:artist_stats}
\end{figure}

\myparagraph{Distribution of real artist images.}
Figure~\ref{fig:artist_stats} shows how frequently each artist's name appears in the captions of our LAION dataset of 100 artists. We compare the frequencies of public-domain artists and non-public-domain artists, using the guideline that an artist's maximum possible U.S. copyright term is 95 years post-mortem~\cite{artistsrights2024}. We find that 38.7\% of the images come from 45 non-public domain artists. As seen in Figure~\ref{fig:artist_stats}, public and non-public domain artists have similar frequency distributions. We use this dataset as real image references for artists' styles for the prototypical network.

\subsection{Prompting Recent Text-to-Image Models with Artist Names}
\label{sec:other_t2i_models}

\begin{figure*}[t]
    \centering
    \includegraphics[width=\linewidth]{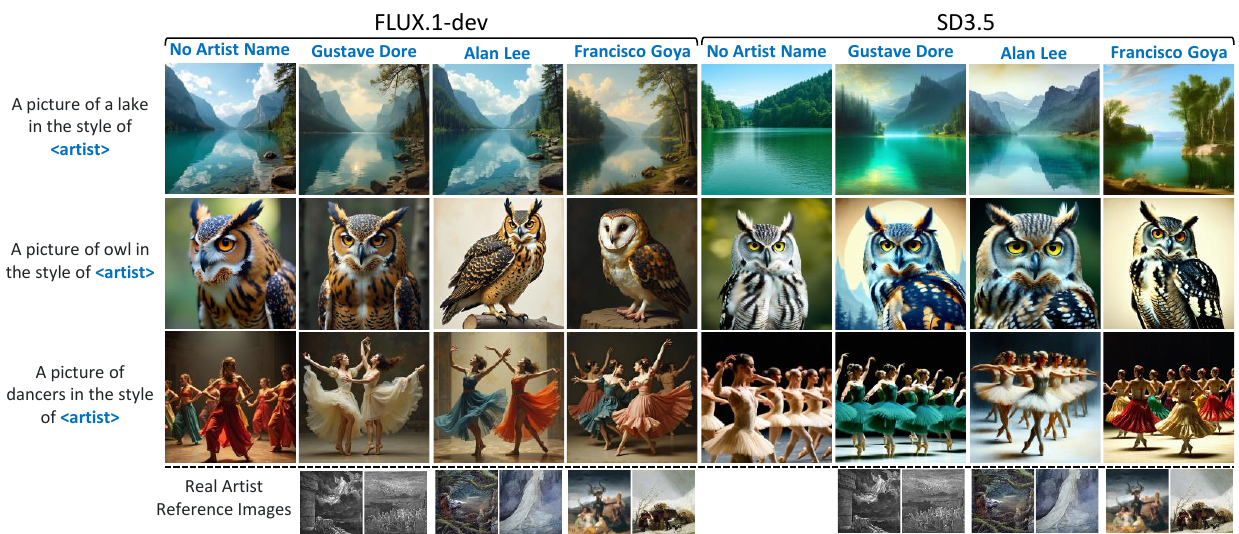}
    \caption{\textbf{FLUX and SD3.5 examples.} 
    For FLUX.1-dev~\cite{flux2024} and SD3.5~\cite{esser2024scaling}, inserting artist names into the prompt frequently has little to no influence on the generated image's style compared to using the same prompt without any artist name. Most of the resulting images are photorealistic, and the generated images do not match the artists' actual styles (bottom row).
    }
    \label{fig:flux_sd35}
\end{figure*}

We tried prompting other recent text-to-image models with artist names, specifically Stable Diffusion 3.5 Large (SD3.5)~\cite{esser2024scaling}, and FLUX.1-dev~\cite{flux2024}, but found that they using artist names often has little to no effect on the generated image style, even when the prompt is simple. For instance, in Figure~\ref{fig:flux_sd35}, we generate images using simple prompts and training set artists from our dataset. Default parameters were used for each model: SD3.5 with 50 inference steps, guidance scale of 7.0, and FLUX.1-dev with 50 inference steps, guidance scale of 3.5. For each prompt, the models generate artist-prompted images that are nearly the same photorealistic style as the images prompted with no artist names. Furthermore, the artist-prompted images are not aligned with the artists' actual styles. 
Since these models often fail to stylize images when prompted with artist names, we do not include these models in our main evaluations.

\subsection{Dataset Statistics Tables}
\label{sec:dataset_statistics_tables}

Following the dataset curation procedure, we create a structured dataset of 1.95M images in order to evaluate vision methods across different axes of generalization. 
The single-artist datasets are summarized in Table~\ref{tbl:dataset_statistics}.
For each artist and prompt combination, we generate images with 2 seeds using the open-weight models SDXL~\cite{podell2023sdxl}, SD1.5~\cite{rombach2022high}, and PixArt-$\Sigma$~\cite{chen2024pixart}. For SDXL images generated with complex prompts, we use 10 different generation seeds. 
To collect images from Midjourney~\cite{midjourney}, a closed-source model, we filter the JourneyDB dataset~\cite{sun2024journeydb} for our set of seen and held-out artists. 
The multi-artist datasets are summarized in Table~\ref{tbl:multi_artist_dataset}. We sample 100 random seen artist combinations and all possible held-out artist combinations for 2 artists and 3 artists, and generate images with SDXL.

\begin{table*}[!h]
\centering
\footnotesize %
    \begin{subtable}[h]{\textwidth}
    \centering
    \caption{\centering \textbf{SDXL}}
    \resizebox{\textwidth}{!}{
        \begin{tabular}{lccccccccc}
        \toprule
    & \multicolumn{4}{c}{Complex Prompts} & \multicolumn{4}{c}{Simple Prompts} & \multicolumn{1}{l}{\multirow{3}{*}{\shortstack{Total\\Unique}}} \\
    \cmidrule(lr){2-5} \cmidrule(lr){6-9}
                                    & \multicolumn{2}{c}{Seen artists}                                   & \multicolumn{2}{c}{Held-out artists}                            & \multicolumn{2}{c}{Seen artists}                                   & \multicolumn{2}{c}{Held-out artists}                            & \\ \cmidrule(lr){2-3} \cmidrule(lr){4-5} \cmidrule(lr){6-7} \cmidrule(lr){8-9}
                                    & \multicolumn{1}{c}{Train} & \multicolumn{1}{c}{Test} & \multicolumn{1}{c}{Test-Time Ref.} & \multicolumn{1}{c}{Test} & \multicolumn{1}{c}{Train} & \multicolumn{1}{c}{Test} & \multicolumn{1}{c}{Test-Time Ref.} & \multicolumn{1}{c}{Test} & \multicolumn{1}{l}{}                              \\ \midrule
        \# Artists                      & 100                                         & 100                      & 10                                   & 10                       & 100                                         & 100                      & 10                                   & 10                       & 110                                               \\
        \# Prompts                      & 900                                         & 100                      & 5                                    & 90                       & 450& 50& 5                                    & 45& 1,500\\
        \# Seeds & 10& 10& 10& 10& 2                                           & 2                        & 2                                    & 2                        & 10                                                \\ \cdashline{1-10}
        \# Images                       & 900,000& 100,000& 500& 9,000& 90,000& 10,000& 100                                  & 900& 1,110,500\\ \bottomrule
        \end{tabular}
    }
    \label{tbl:our_dataset}
    \end{subtable}

    \begin{subtable}[h]{0.65\textwidth}
    \centering
    \caption{\centering \textbf{SD1.5, PixArt}}
    \resizebox{\textwidth}{!}{
        \begin{tabular}{lccccccccc}
        \toprule
    & \multicolumn{4}{c}{Complex Prompts} & \multicolumn{4}{c}{Simple Prompts} & \multicolumn{1}{l}{\multirow{3}{*}{\shortstack{Total\\Unique}}} \\
    \cmidrule(lr){2-5} \cmidrule(lr){6-9}
                                    & \multicolumn{2}{c}{Seen artists}                                   & \multicolumn{2}{c}{Held-out artists}                            & \multicolumn{2}{c}{Seen artists}                                   & \multicolumn{2}{c}{Held-out artists}                            & \\ \cmidrule(lr){2-3} \cmidrule(lr){4-5} \cmidrule(lr){6-7} \cmidrule(lr){8-9}
                                    & \multicolumn{1}{c}{Train} & \multicolumn{1}{c}{Test} & \multicolumn{1}{c}{Test-Time Ref.} & \multicolumn{1}{c}{Test} & \multicolumn{1}{c}{Train} & \multicolumn{1}{c}{Test} & \multicolumn{1}{c}{Test-Time Ref.} & \multicolumn{1}{c}{Test} & \multicolumn{1}{l}{}                              \\ \midrule
        \# Artists                      & 100                                         & 100                      & 10                                   & 10                       & 100                                         & 100                      & 10                                   & 10                       & 110                                               \\
        \# Prompts                      & 450& 50& 5                                    & 45& 450& 50& 5                                    & 45& 1,000\\
        \# Seeds & 2& 2& 2& 2& 2                                           & 2                        & 2                                    & 2                        & 2\\ \cdashline{1-10}
        \# Images                       & 90,000& 10,000& 100& 900& 90,000& 10,000& 100                                  & 900& 202,000\\ \bottomrule
        \end{tabular}
    }
    \label{tbl:pixart_sd15_dataset}
    \end{subtable}%
    \hspace{2mm}
    \begin{subtable}[h]{0.33\textwidth}
    \centering
    \caption{\centering \textbf{Midjourney}}
    \resizebox{\textwidth}{!}{
        \begin{tabular}{lccccc}
        \toprule
        & \multicolumn{4}{c}{Complex Prompts} & \multicolumn{1}{c}{\multirow{3}{*}{\shortstack{Total\\Unique}}} \\
        \cmidrule(lr){2-5}
        & \multicolumn{2}{c}{Seen artists} & \multicolumn{2}{c}{Held-out artists} & \\
        \cmidrule(lr){2-3} \cmidrule(lr){4-5}
        & \multicolumn{1}{c}{Train} & \multicolumn{1}{c}{Test} & \multicolumn{1}{c}{Test-Time Ref.} & \multicolumn{1}{c}{Test} & \\ \midrule
        \# Artists & 35 & 35 & 95 & 95 & 130 \\
        \# Prompts & 647 & 658 & 649 & 697 & 2,651 \\ \cdashline{1-6}
        \# Images  & 647 & 658 & 649 & 697 & 2,651 \\ 
        \bottomrule
        \end{tabular}
    }
    \label{tbl:midjourney_dataset}
    \end{subtable}

    \caption{\textbf{Single-artist dataset statistics.} 
    The prompted artist identification dataset includes images generated with SDXL~\cite{podell2023sdxl}, SD1.5~\cite{rombach2022high}, PixArt-$\Sigma$~\cite{chen2024pixart}, and Midjourney~\cite{midjourney}. As described in Section~\ref{sec:dataset}, we collect 1,000 complex prompts, 500 simple prompts, and use 110 different artist names in the prompts. For each artist and prompt combination, we generate images using 10 different seeds for SDXL complex prompts and 2 different seeds for the other subsets. We split the 110 artists into 100 seen and 10 held-out artists. For seen artists, we use a separate set of prompts for testing. For held-out artists, we additionally split test prompt images for test-time reference. 
    }
    \label{tbl:dataset_statistics}
\end{table*}

\begin{table*}[!h]
\centering
    \begin{subtable}[h]{\textwidth}
    \centering
    \caption{\textbf{2 Artists}}
    \resizebox{\linewidth}{!}{
    \begin{tabular}{lccccccccc}
    \toprule
    & \multicolumn{4}{c}{Complex Prompts} & \multicolumn{4}{c}{Simple Prompts} & \multirow{3}{*}{\shortstack{Total\\Unique}} \\
    \cmidrule(lr){2-5} \cmidrule(lr){6-9}
    & \multicolumn{2}{c}{Seen artists} & \multicolumn{2}{c}{Held-out artists} & \multicolumn{2}{c}{Seen artists} & \multicolumn{2}{c}{Held-out artists} & \\
    \cmidrule(lr){2-3} \cmidrule(lr){4-5} \cmidrule(lr){6-7} \cmidrule(lr){8-9}
    & Train & Test & Test-Time Ref. & Test & Train & Test & Test-Time Ref. & Test & \\
    \midrule
    \# Artists & 100 & 100 & 10 & 10 & 100 & 100 & 10 & 10 & 110 \\
    \# Prompts & 450 & 50 & 5 & 45 & 450 & 50 & 5 & 45 & 1,000 \\
    \# Seeds & 2 & 2 & 2 & 2 & 2 & 2 & 2 & 2 & 2 \\
    \# 2-artist combinations & 100 & 100 & 45 & 45 & 100 & 100 & 45 & 45 & 145 \\
    \cdashline{1-10}
    \# Images & 90,000 & 10,000 & 450 & 4,050 & 90,000 & 10,000 & 450 & 4,050 & 209,000 \\
    \bottomrule
    \end{tabular}
    }
    \label{tbl:two_artist_dataset}
    \end{subtable}

    \begin{subtable}[h]{\textwidth}
    \centering
    \caption{\textbf{3 Artists}}
    \resizebox{\linewidth}{!}{
    \begin{tabular}{lccccccccc}
    \toprule
    & \multicolumn{4}{c}{Complex Prompts} & \multicolumn{4}{c}{Simple Prompts} & \multirow{3}{*}{\shortstack{Total\\Unique}} \\
    \cmidrule(lr){2-5} \cmidrule(lr){6-9}
    & \multicolumn{2}{c}{Seen artists} & \multicolumn{2}{c}{Held-out artists} & \multicolumn{2}{c}{Seen artists} & \multicolumn{2}{c}{Held-out artists} & \\
    \cmidrule(lr){2-3} \cmidrule(lr){4-5} \cmidrule(lr){6-7} \cmidrule(lr){8-9}
    & Train & Test & Test-Time Ref. & Test & Train & Test & Test-Time Ref. & Test & \\
    \midrule
    \# Artists & 100 & 100 & 10 & 10 & 100 & 100 & 10 & 10 & 110 \\
    \# Prompts & 450 & 50 & 5 & 45 & 450 & 50 & 5 & 45 & 1,000 \\
    \# Seeds & 2 & 2 & 2 & 2 & 2 & 2 & 2 & 2 & 2 \\
    \# 3-artist combinations & 100 & 100 & 120 & 120 & 100 & 100 & 120 & 120 & 220 \\
    \cdashline{1-10}
    \# Images & 90,000 & 10,000 & 1,200 & 10,800 & 90,000 & 10,000 & 1,200 & 10,800 & 224,000 \\
    \bottomrule
    \end{tabular}
    }
    \label{tbl:three_artist_dataset}
    \end{subtable}
\vspace{-2mm}
\caption{\textbf{Multi-artist dataset statistics.} To evaluate prompted artist identification of images generated with multiple artists in the prompt, we generate datasets of images prompted with 2 artists and 3 artists. The datasets are generated using the same procedure as the single-artist dataset, except we sample 100 artist combinations for each number of artists.
}
\label{tbl:multi_artist_dataset}
\end{table*}

\section{Training Details}
\label{sec:training_details}

\myparagraph{Prototypical Network.}
We train the prototypical network~\cite{snell2017prototypical} for prompted artist identification by fine-tuning the pre-trained CLIP~\cite{radford2021learning} ViT-L/14 image encoder end-to-end on our prompted artist identification dataset.
We apply blur and JPEG training data augmentation by using the same recommended setting as Wang~\etal~\cite{Wang_2020_cnndet}: the image is randomly blurred and JPEG-ed, each with $10\%$ probability. Random resized crop and horizontal flip are also applied.
We sweep learning rates, $1e-7$, $1e-6$, and $1e-5$, and number of training epochs, 1 through 5.
We choose the best-performing configuration: fine-tune for 1 epoch with a learning rate of $1e-6$, temperature $\tau$ of 0.07, and a batch size of 128 per GPU, distributed across 4 GPUs. This leads to an effective batch size of 512. The compute environment is 4 A6000 GPUs with 40GB of vRAM each and mixed precision (fp16) training.

\myparagraph{Vanilla Classifier.}
We add an MLP with one hidden layer as the classification head to the pre-trained CLIP~\cite{radford2021learning} ViT-L/14 image encoder, then train all layers on our dataset. 
The training settings are consistent with the prototypical network: fine-tune for 1 epoch with a learning rate of $1e-6$, effective batch size of 512, and fp16 mixed precision training.

\section{Experiments and Ablations}
\label{sec:method_exps_ablations}
We analyze and validate the design choices for each of the visual representation methods we evaluate on our benchmark in the main paper.
All experiments were conducted with SDXL images from our prompted artist identification dataset. All complex prompts, and a subset of 100 simple prompts (90 train, 10 test), were included in the dataset.

\subsection{Retrieve from Real Images}
\label{sec:retrieve_real}
Retrieval-based methods can use either real artist images or generated images as the reference database to classify artist-prompted images. 
We compare both settings on our benchmark and find that for each method, retrieval from generated images outperforms retrieval from real images on most evaluation sets, and the final performance ranking of the methods remains largely unchanged. We use real artist images collected from LAION-Styles, the same set of images used to compute the prototypes for the prototypical network. The numbers of images for seen and held-out artists is shown in Table~\ref{tbl:laion_dataset}.
For retrieval from generated images, we use all of the generated seen artist training images from our SDXL, SD1.5, PixArt, and Midjourney datasets, including both simple and complex-prompted images. 
In Table~\ref{tbl:sdxl_eval}, we report results on SDXL images and observe that retrieval from real images leads to lower performance than retrieval from generated images for all methods, across all evaluation sets except for images with complex prompts and held-out artists. Thus, we focus on retrieval from generated images in our main benchmark evaluation.

\begin{table}[h]
\centering
\resizebox{0.4\textwidth}{!}{
\begin{tabular}{lccc}
\toprule
        & Seen artists          & Held-out artists        & Total \\ \midrule
\# Artists & 100 & 10  & 110    \\ \cdashline{1-4}
\# Images   & 9,907 & 860 & 10,767 \\
\bottomrule
\end{tabular}
}
\caption{\textbf{Real artist image dataset}. We use a subset of LAION-Styles~\cite{somepalli2024measuringstyle} containing 100 seen artists and 10 held-out artists. 
Then, we benchmark prompted artist identification of various visual representation methods and use this dataset for real image references of each artist's style.
}
\label{tbl:laion_dataset}
    \vspace{-3mm}
\end{table}

\subsection{Retrieve from Prototypes}
\label{sec:retrieve_prototype}
In the main benchmark evaluation, we evaluate retrieval-based methods that use individual reference image embeddings for retrieval. However, the prototypical network learns classification via prototype alignment, where we set an artist's prototype to be the averaged feature of the real reference images of the artist's work. 

For a fair comparison, we test retrieval-based methods in a similar setting by retrieving from a set of averaged embeddings of real images, computing one average embedding per artist (same as the prototype). The difference is that the methods are \textit{not} fine-tuned to align with the prototypes. We evaluate this average embedding retrieval baseline for CSD~\cite{somepalli2024measuringstyle} and CLIP~\cite{radford2021learning} and compare the results to the prototypical network in Table~\ref{tab:nn_retrieve_avg}. We find that the prototypical network outperforms them in most categories.

\begin{table}[h]
\centering
\resizebox{\linewidth}{!}{%
\begin{tabular}{lcllll}
\toprule
\multicolumn{1}{c}{\multirow{3}{*}{Method}} & \multirow{3}{*}{\shortstack[c]{Reference/\\Prototype}} & \multicolumn{4}{c}{Evaluation Set}                                                                                  \\ \cmidrule(lr){3-6} 
\multicolumn{1}{c}{}                        &                                      & \multicolumn{2}{c}{Seen artists (100-way)}               & \multicolumn{2}{c}{Held-out (10-way)}                    \\ \cmidrule(lr){3-4} \cmidrule(lr){5-6} 
\multicolumn{1}{c}{}                        &                                      & \multicolumn{1}{c}{Complex} & \multicolumn{1}{c}{Simple} & \multicolumn{1}{c}{Complex} & \multicolumn{1}{c}{Simple} \\ \midrule
{CLIP~\cite{radford2021learning}}& Artist Avg.               & 15.1                        & 38.3                       & 44.9                        & 79.0                       \\
{CSD~\cite{somepalli2024measuringstyle}} & Artist Avg.               & 19.7                        & 48.1                       & 56.8                        & \textbf{92.0}              \\
Proto. Net.~\cite{snell2017prototypical}                            & Artist Avg.               & \textbf{43.2}               & \textbf{87.0}              & \textbf{62.7}               & 87.0                       \\ \bottomrule
\end{tabular}%
}
\caption{\textbf{Comparison to baselines retrieving from prototypes}. 
We additionally evaluate average embedding retrieval for baselines CSD and CLIP and compare them with the prototypical network.
}
\label{tab:nn_retrieve_avg}
\end{table}

\subsection{k-NN Evaluation Results}
\label{sec:knn_eval}
In our main evaluations of retrieval-based methods on our benchmark, we use $k$-NN classification with $k=1$, as we find that increasing $k$ does not help. 

\begin{figure}[!h]
    \centering
    \resizebox{\linewidth}{!}{
    \includegraphics[width=\linewidth]{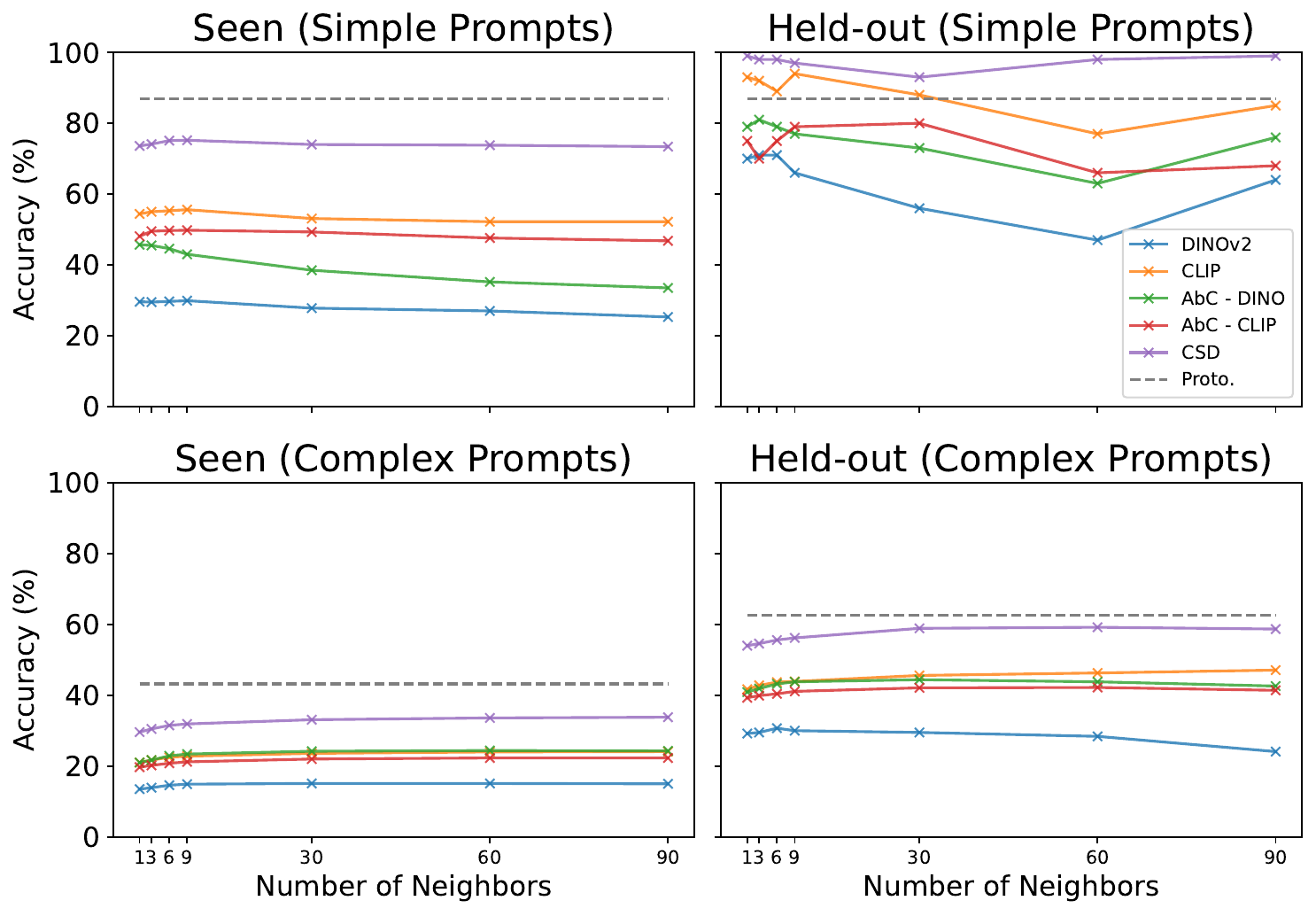}
    }
    \caption{\textbf{Effect of $k$ on accuracy for $k$-NN classification}. 
    We perform $k$-NN evaluation on multiple $k$ values and find that for each retrieval-based method, the performance only changes a small amount as $k$ increases. Thus, the performance ranking of the methods remains unchanged across different $k$ values and the gap between our method and baselines remains the same.
    }
\label{fig:knn_results}
\end{figure}

We test $k$-NN classification with majority voting, where the final predicted class label is the majority class label within the set of the top $k$ nearest neighbors, and ties are broken by the majority class label with the closest retrieved embedding to the query embedding. 
We evaluate multiple $k$ values for $k$-NN majority voting classification to investigate the effect of $k$ on artist classification accuracy in Figure~\ref{fig:knn_results}. 

First, we find that the performance ranking of the methods remains unchanged across different $k$ values. Also, for each retrieval-based method, the performance only changes a small amount as $k$ increases. The exception to these trends is when the images are generated with held-out artists and simple prompts, where the performance saturates at a very small $k$.

\vspace{-5mm}
\subsection{Classification Method Experiments}
\label{sec:classifier_exps}
We analyze and validate the design choices for the classification methods we evaluate on our benchmark in the main paper. Table~\ref{tbl:ablation} shows results on SDXL images from our prompted artist identification dataset, averaged across the complex and simple prompt evaluation sets.

\myparagraph{Comparing classifier types.}
In Table~\ref{tab:abl_model_type}, we train and compare different model types: vanilla classification, supervised contrastive learning~\cite{khosla2020supervised}, and prototypical networks~\cite{snell2017prototypical}. 
While the classifier outperforms prototypical networks on seen artist classification, by default, it cannot classify held-out artists because of the fixed label set. To test generalization, we remove the final linear layer of the classifier to extract features for retrieval. We find that retrieving from real or generated images performs worse than the prototypical network in held-out artist classification. This indicates training with prototypes of reference images helps with generalization.

Next, we try metric learning by running Supervised Contrastive Learning~\cite{khosla2020supervised} on the generated images, with all images from the same artist using the same label. We find that this does not achieve strong performance and focus on prototypical networks and vanilla classification for benchmark evaluation.

\myparagraph{Nearest neighbors and different sources for prototypes.}
Prototypical networks can also be used as a feature extractor to run retrieval, which performs on par with using the prototype for classification. We report results Table~\ref{tab:abl_nn}. In addition, using generated images as prototypes leads to worse performance, so we use real images as prototypes in benchmark evaluations. We hypothesize that
using a different modality (e.g., real artist images) as prototypes encourages the encoder to extract artist-specific features more, helping overall generalization.

\myparagraph{Which training data helps?}
We vary the training set of the prototypical network and report performance in Table~\ref{tab:abl_train_set}. First, instead of training to classify artists from generated images, we train a prototypical network using real artist images as input. Training a model without synthesized images drastically hurts performance. Since good classification performance requires extracting artist-related cues even when images are generated from complex prompts, training an encoder from generated images can help learn such cues.

We further confirm this hypothesis by training a prototypical network using only generated images with easy prompts. While this variant performs better than the one trained only with real images, the performance remains worse than the ones trained on generated images with complex prompts. On the other hand, training only with complex prompts yields similar performance as training on the dataset with simple and complex prompts combined.

\begin{table*}[ht]
\centering
\footnotesize %
    \hskip-0.8cm
    \begin{subtable}[t]{0.31\textwidth}
    \centering
    \caption{\centering \textbf{Model type}}
    \resizebox{1.1\textwidth}{!}{%
    \begin{tabular}{llcc}
    \toprule
     \multicolumn{2}{c}{Model} &  \multicolumn{2}{c}{Evaluation set} \\ \cmidrule(lr){1-2} \cmidrule(lr){3-4}
     Training & Testing &  \shortstack[c]{Seen Artists\\ (100-way)} & \shortstack[c]{Held-out \\ (10-way)} \\ \midrule
    \multirow{3}{*}{\shortstack[c]{Vanilla\\Classifier}} & Classifier &  \textbf{53.4}& -- \\
     & NN - Real &  24.4& 45.0\\
     & NN - Gen. &  51.3& 58.0\\ \cdashline{1-4}
    \multirow{3}{*}{\shortstack[l]{Supervised\\Contrastive\\Learning~\cite{khosla2020supervised}}} & Classifier &  49.3 & -- \\
     & NN - Real  & 24.4& 21.6\\
     & NN - Gen. & 47.5& 29.6\\ \midrule
    Proto. Net.~\cite{snell2017prototypical} & Proto. - Real &  43.9 & \textbf{63.0}\\
    \bottomrule
    \end{tabular}
    }
    \label{tab:abl_model_type}
    \end{subtable}%
\hspace{5mm}
    \begin{subtable}[t]{0.31\textwidth}
    \centering
    \caption{\centering \textbf{Training set}}
    \resizebox{1\textwidth}{!}{%
    \begin{tabular}{lcc}
    \toprule
     \multirow{2}{*}{Model} &  \multicolumn{2}{c}{Evaluation set} \\ \cmidrule(lr){2-3} 
     & \shortstack[c]{Seen artists \\ (100-way)} & \shortstack[c]{Held-out \\ (10-way)} \\ \midrule
    Real  & 16.5 & 49.7 \\
    Simple only  & 21.1 & 55.2 \\
    Complex only  & 43.6 & 62.5 \\ \midrule
    Simple + Complex &  \textbf{43.9}& \textbf{63.0}\\
    \bottomrule
    \end{tabular}
    }
    \label{tab:abl_train_set}
    \end{subtable}%
\hspace{1mm}
    \begin{subtable}[t]{0.31\textwidth}
    \centering
    \caption{{\centering \textbf{Proto. vs. Nearest Neighbors}}}
    \vspace{0.2mm}
    \resizebox{1.2\textwidth}{!}{%
    \begin{tabular}{lcccc}
    \toprule
     & \multicolumn{2}{c}{Reference} & \multicolumn{2}{c}{Evaluation set} \\ \cmidrule(lr){2-3} \cmidrule(lr){4-5}
    Method & Real & Gen. & \shortstack[c]{Seen artists \\ (100-way)} & \shortstack[c]{Held-out \\ (10-way)} \\ \midrule
    NN - Real & \checkmark & & 26.7 & 54.3 \\
    NN - Gen. & & \checkmark & \textbf{43.9}& 62.3 \\
    Proto. - Gen. & & \checkmark & 23.0 & 61.2 \\ \midrule
    Proto. - Real & \checkmark & & \textbf{43.9}& \textbf{63.0}\\
    \bottomrule
    \end{tabular}
    }
    \label{tab:abl_nn}
    \end{subtable}
\caption{\textbf{Classification method experiments}. 
In the main paper, we benchmark two classification methods: prototypical networks and vanilla classification. Here, we test more classifier method variants and training datasets, reporting accuracy numbers averaged across the SDXL complex and simple prompt evaluation sets. 
(a) We test {\bf model types}, comparing vanilla classification, supervised contrastive learning, and prototypical networks.
As classifiers have a fixed label set, we use nearest neighbors on their feature space when testing on held-out artists. 
Supervised contrastive learning performs worse than both the prototypical network and vanilla classification, so we do not include it in our benchmark evaluation.
(b) We test {\bf training sets} for the prototypical network. Learning from real images is not as effective as learning from generated images. Using both simple and complex prompts outperforms using one or the other, demonstrating they provide complementary information. 
(c) For the prototypical network, we compare using the {\bf prototype vs. nearest neighbors}.
When using the learned prototypical network representation for nearest neighbors, performance slightly degrades. Using the generated images for prototypes, rather than real, also degrades performance, so we use real images as prototypes for our main implementation.
}
\label{tbl:ablation}
\end{table*}

\section{Evaluation Details}
\label{sec:eval_details}

\subsection{Bootstrapping Procedure}
\begin{figure}[!ht]
    \centering
    \resizebox{\linewidth}{!}{
    \includegraphics[width=\linewidth]{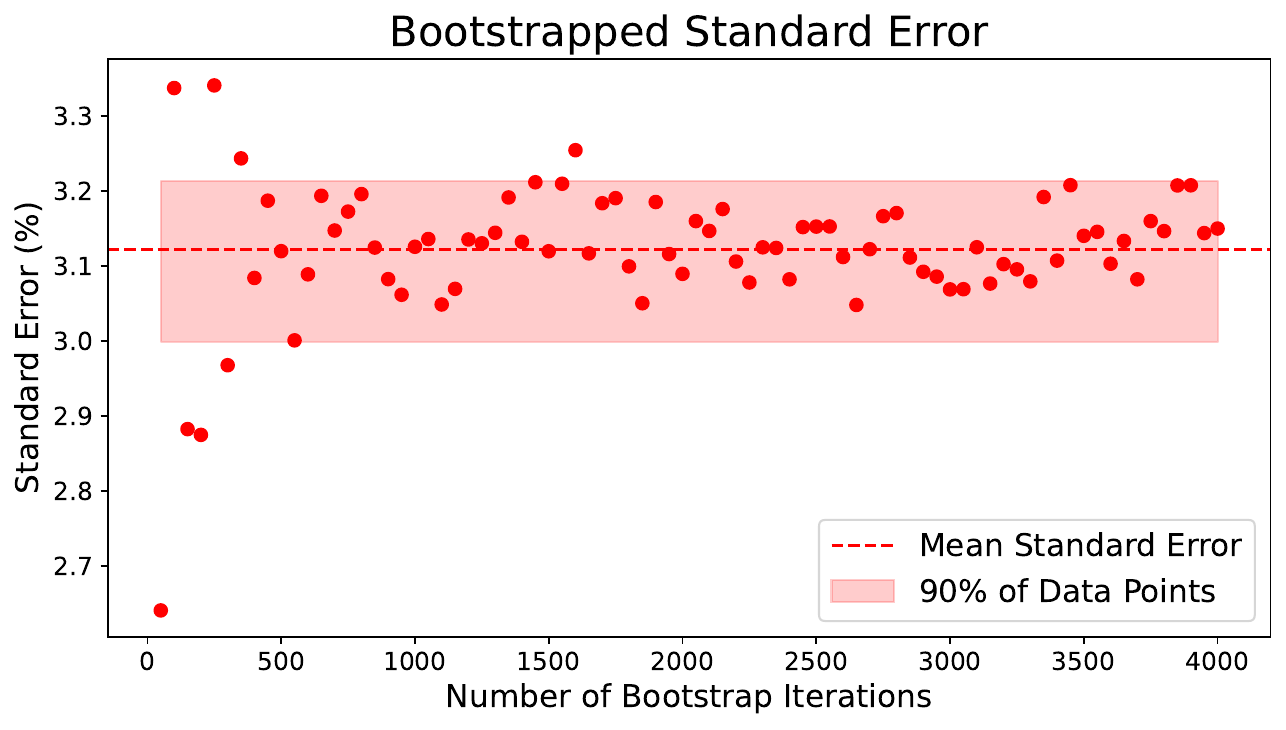}
    }
    \caption{\textbf{Bootstrapping procedure convergence}. 
    We plot standard error against the number of bootstrap iterations for the prototypical network's artist classification accuracy on SDXL images and simple prompts. 
    We use 2000 bootstrap iterations in our main benchmark evaluation because standard error converges at around 2000 iterations, where the standard error remains within the range of 90\% of the data points.
    }
    \label{fig:bootstrap_convergence}
\end{figure}

 To estimate the statistical significance of each evaluation on our benchmark, we use a bootstrapping procedure. We bootstrap each model's predictions by resampling the evaluation images, with replacement, by artist name, prompt template, and generation seed for 2000 iterations.
 We selected 2000 iterations by plotting the convergence of the standard error, as shown in Figure~\ref{fig:bootstrap_convergence}. 

 Given the prototypical network's artist classification predictions on SDXL images and simple prompts, we plot the standard error of the classification accuracy against the number of bootstrap iterations. The number of bootstrapping iterations was increased from 50 to 4000 in increments of 50. The standard error remains within the range of 90\% of the data points at around 2000 iterations, indicating that the standard error converges at this point. Thus, we use 2000 bootstrap iterations in our main benchmark evaluation.

\subsection{Quantitative Result Tables}
We include the quantitative results from our main benchmark evaluation in Section~\ref{sec:results} as tables below. 
The single-artist evaluation results include SDXL images in Table~\ref{tbl:sdxl_eval}, SD1.5 images in Table~\ref{tbl:sd15_eval}, PixArt-$\Sigma$ images in Table~\ref{tbl:pixart_eval}, and Midjourney images in Table~\ref{tbl:midjourney_eval}.
The multi-artist evaluation results include 2-artist prompted images in Table~\ref{tbl:two_artist_eval} and 3-artist prompted images in Table~\ref{tbl:three_artist_eval}.

\begin{table*}[!t]
\centering
\resizebox{\linewidth}{!}{
\begin{tabular}{llcccccccc} \toprule
 & & & \multicolumn{2}{c}{\multirow{2}{*}{\raisebox{-.75\height}{\shortstack[c]{Reference/ \\ Prototype images}}}} & \multicolumn{5}{c}{Evaluation set} \\ \cmidrule(lr){6-10}

 & & & & & \multicolumn{2}{c}{Seen artists (100-way)} & \multicolumn{2}{c}{Held-out (10-way)} \\ \cmidrule(lr){4-5} \cmidrule(lr){6-7} \cmidrule(lr){8-10}
 Family & Method & Architecture & Real & Gen. & \shortstack[c]{Complex} & \shortstack[c]{Simple} & \shortstack[c]{Complex} & \shortstack[c]{Simple} \\ \midrule
Chance & -- & -- & -- & -- & 1.0 & 1.0 & 10.0 & 10.0 \\ \midrule

\multirow{10}{*}{\shortstack[l]{Retrieval-\\based\\ methods}} 
& \multirow{2}{*}{AbC - DINO~\cite{wang2023evaluating}} & \multirow{2}{*}{ViT-B/16}
    & \checkmark & & 7.2 ± 1.0 & 13.9 ± 2.0 & 33.9 ± 6.4 & 47.7 ± 9.4 \\
& & & & \checkmark & 21.7 ± 2.4 & 45.1 ± 3.7 & 30.7 ± 5.5 & 55.8 ± 7.5 \\ \cdashline{2-10}

& \multirow{2}{*}{AbC - CLIP~\cite{wang2023evaluating}} & \multirow{2}{*}{ViT-B/16}
    & \checkmark & & 10.5 ± 1.5 & 20.4 ± 2.7 & 35.4 ± 6.3 & 52.3 ± 8.4 \\
& & & & \checkmark & 20.3 ± 2.4 & 48.1 ± 3.4 & 30.2 ± 4.0 & 60.0 ± 5.9 \\ \cdashline{2-10}

& \multirow{2}{*}{DINOv2~\cite{oquab2023dinov2}} & \multirow{2}{*}{ViT-L/14}
    & \checkmark & & 6.0 ± 1.0 & 11.0 ± 1.9 & 29.6 ± 5.5 & 44.6 ± 8.2 \\
& & & & \checkmark & 14.1 ± 1.9 & 28.7 ± 3.3 & 16.4 ± 3.4 & 37.2 ± 6.8 \\ \cdashline{2-10}

& \multirow{2}{*}{CLIP~\cite{radford2021learning}} & \multirow{2}{*}{ViT-L/14}
    & \checkmark & & 12.4 ± 1.7 & 24.9 ± 2.9 & 41.1 ± 6.8 & 62.7 ± 8.3 \\
& & & & \checkmark & 22.8 ± 2.6 & 53.4 ± 3.5 & 35.3 ± 4.2 & 78.5 ± 4.7 \\ \cdashline{2-10}

& \multirow{2}{*}{CSD~\cite{somepalli2024measuringstyle}} & \multirow{2}{*}{ViT-L/14}
    & \checkmark & & 15.3 ± 2.0 & 31.9 ± 3.3 & 47.4 ± 7.0 & 74.3 ± 6.6 \\ 
& & & & \checkmark & 30.1 ± 2.9 & 66.4 ± 3.4 & 44.4 ± 4.8 & \textbf{84.6 ± 3.9} \\ \midrule

\multirow{1}{*}{{\shortstack[c]{Fine-tuned\\classifiers}}} & Prototypical Network~\cite{snell2017prototypical} & ViT-L/14
    & \checkmark & & 40.3 ± 3.1 & 74.9 ± 3.0 & \textbf{57.7 ± 4.5} & 75.5 ± 7.9  \\ \cdashline{2-10}

 & Vanilla Classifier& ViT-L/14 & & \checkmark & \textbf{40.7 ± 3.2}  & \textbf{76.4 ± 3.0} & -- & -- \\ \bottomrule
\end{tabular}
}

\caption{\textbf{Classification accuracy on SDXL images}. We compare retrieval-based methods and fine-tuned classifiers on SDXL images from our prompted artist identification dataset. While retrieval-based methods can use either real or generated images as the reference database, we find that retrieval from generated images outperforms retrieval from real images for all methods, across all evaluation sets except for images with complex prompts and held-out artists. Thus, we focus on retrieval from generated images in our main benchmark evaluation.
}
\vspace{-2mm}
\label{tbl:sdxl_eval}
\end{table*}

\begin{table*}[!t]
\centering
\resizebox{0.7\linewidth}{!}{
\begin{tabular}{llcccccc} \toprule
 & & \multicolumn{5}{c}{Evaluation set} \\ \cmidrule(lr){3-8}

 & & \multicolumn{2}{c}{Seen artists (100-way)} & \multicolumn{2}{c}{Held-out (10-way)} \\ \cmidrule(lr){3-4} \cmidrule(lr){5-6}
 Family & Method & \shortstack[c]{Complex} & \shortstack[c]{Simple} & \shortstack[c]{Complex} & \shortstack[c]{Simple} \\ \midrule
Chance & -- & 1.0 & 1.0 & 10.0 & 10.0 \\ \midrule

\multirow{5}{*}{\shortstack[l]{Retrieval-\\based\\ methods}} 
& AbC - DINO~\cite{wang2023evaluating} & 26.5 ± 3.3 & 49.2 ± 4.1 & 27.5 ± 6.1 & 51.8 ± 8.0 \\ 

& AbC - CLIP~\cite{wang2023evaluating} & 24.4 ± 3.2 & 48.9 ± 3.9 & 29.2 ± 6.1 & 50.6 ± 8.5 \\ 

& DINOv2~\cite{oquab2023dinov2} & 18.8 ± 2.7 & 36.3 ± 3.7 & 22.7 ± 5.4 & 41.5 ± 7.6 \\ 

& CLIP~\cite{radford2021learning} & 27.3 ± 3.5 & 52.9 ± 3.8 & 31.5 ± 6.2 & 57.5 ± 8.3 \\ 

& CSD~\cite{somepalli2024measuringstyle} & 32.1 ± 3.7 & 61.3 ± 3.8 & 34.6 ± 6.2 & \textbf{66.3 ± 7.1} \\ \midrule

\multirow{1}{*}{{\shortstack[c]{Fine-tuned\\classifiers}}} 
& Prototypical Network~\cite{snell2017prototypical} & \textbf{40.5 ± 3.8} & 63.0 ± 3.7 & \textbf{43.8 ± 6.4} & 53.8 ± 8.7  \\ 

& Vanilla Classifier& 40.2 ± 3.8  & \textbf{65.2 ± 3.7} & -- & -- \\ \bottomrule
\end{tabular}
}
\caption{\textbf{Classification accuracy on SD1.5 images}.}
\label{tbl:sd15_eval}
\end{table*}	

\begin{table*}[!t]
\centering
\resizebox{0.7\linewidth}{!}{
\begin{tabular}{llcccccc} \toprule
 & & \multicolumn{5}{c}{Evaluation set} \\ \cmidrule(lr){3-7}

 & & \multicolumn{2}{c}{Seen artists (100-way)} & \multicolumn{2}{c}{Held-out (10-way)} \\ \cmidrule(lr){3-4} \cmidrule(lr){5-6}
 Family & Method & \shortstack[c]{Complex} & \shortstack[c]{Simple} & \shortstack[c]{Complex} & \shortstack[c]{Simple} \\ \midrule
Chance & -- & 1.0 & 1.0 & 10.0 & 10.0 \\ \midrule
\multirow{5}{*}{\shortstack[l]{Retrieval-\\based\\ methods}}
 & AbC - DINO~\cite{wang2023evaluating} & 6.8 ± 1.3 & 23.3 ± 2.7 & 7.8 ± 2.5 & 20.4 ± 6.6 \\
 & AbC - CLIP~\cite{wang2023evaluating} & 6.6 ± 1.4 & 24.3 ± 2.9 & 6.6 ± 2.4 & 22.6 ± 6.7 \\
 & DINOv2~\cite{oquab2023dinov2} & 4.7 ± 1.0 & 13.0 ± 1.9 & 4.2 ± 1.6 & 9.5 ± 3.5 \\
 & CLIP~\cite{radford2021learning} & 7.4 ± 1.5 & 26.6 ± 2.9 & 9.3 ± 3.1 & 29.6 ± 7.0 \\
 & CSD~\cite{somepalli2024measuringstyle} & 9.4 ± 1.9 & 38.0 ± 3.4 & 8.8 ± 3.0 & \textbf{35.3 ± 7.2} \\ \midrule
\multirow{1}{*}{{\shortstack[c]{Fine-tuned\\classifiers}}}
 & Prototypical Network~\cite{snell2017prototypical} & 12.6 ± 2.0 & 40.8 ± 3.5 & \textbf{14.9 ± 3.3} & 23.8 ± 6.4  \\

 & Vanilla Classifier& \textbf{12.9 ± 2.0}  & \textbf{41.9 ± 3.5} & -- & -- \\ \bottomrule
\end{tabular}
}
\caption{\textbf{Classification accuracy on PixArt-$\Sigma$ images}.}
\label{tbl:pixart_eval}
\end{table*}

\begin{table*}[!t]
\centering
\resizebox{0.6\linewidth}{!}{
\begin{tabular}{llcc} \toprule
 & & \multicolumn{2}{c}{Evaluation set} \\ \cmidrule(lr){3-4}
 & & Seen artists (34-way) & Held-out (96-way) \\ \cmidrule(lr){3-3} \cmidrule(lr){4-4}
 Family & Method & \shortstack[c]{Complex} & \shortstack[c]{Complex} \\ \midrule
Chance & -- & 2.9 & 1.0 \\ \midrule

\multirow{5}{*}{\shortstack[l]{Retrieval-\\based\\ methods}} 
& AbC - DINO~\cite{wang2023evaluating} & 19.8 ± 2.0 & 32.3 ± 2.6 \\
& AbC - CLIP~\cite{wang2023evaluating} & 29.3 ± 2.8 & 39.7 ± 3.2 \\
& DINOv2~\cite{oquab2023dinov2} & 16.8 ± 2.0 & 28.2 ± 2.4 \\
& CLIP~\cite{radford2021learning} & 30.3 ± 3.1 & 42.7 ± 2.9 \\
& CSD~\cite{somepalli2024measuringstyle} & 34.4 ± 2.3 & \textbf{42.8 ± 3.3} \\ \midrule
\multirow{1}{*}{{\shortstack[c]{Fine-tuned\\classifiers}}} 
& Prototypical Network~\cite{snell2017prototypical} & \textbf{56.8 ± 3.2} & 16.8 ± 2.7 \\
& Vanilla Classifier& 57.6 ± 2.7 & -- \\ \bottomrule 
\end{tabular}
}
\caption{\textbf{Classification accuracy on Midjourney images}.}
\label{tbl:midjourney_eval}
\end{table*}

\begin{table*}[!t]
\centering
\resizebox{0.9\linewidth}{!}{
\begin{tabular}{llcccccc} \toprule
 & & \multicolumn{5}{c}{Evaluation set} \\ \cmidrule(lr){3-7}

 & & \multicolumn{2}{c}{Seen artists (100-way)} & \multicolumn{2}{c}{Held-out (10-way)} \\ \cmidrule(lr){3-4} \cmidrule(lr){5-6}
 Family & Method & \shortstack[c]{Complex} & \shortstack[c]{Simple} & \shortstack[c]{Complex} & \shortstack[c]{Simple} \\ \midrule
Chance & -- & 3.0 & 3.0 & 37.2 & 37.2 \\ \midrule
\multirow{5}{*}{\shortstack[l]{Retrieval-\\based\\ methods}}
    & AbC - DINO~\cite{wang2023evaluating} & 15.7 ± 1.6 & 28.2 ± 1.6 & 40.1 ± 1.9 & 33.2 ± 2.3 \\
    & AbC - CLIP~\cite{wang2023evaluating} & 15.2 ± 1.7 & 30.8 ± 1.5 & 39.3 ± 2.1 & 34.7 ± 2.5 \\
    & DINOv2~\cite{oquab2023dinov2} & 11.2 ± 1.3 & 20.1 ± 1.6 & 36.3 ± 1.9 & 33.6 ± 2.2 \\
    & CLIP~\cite{radford2021learning} & 16.5 ± 1.7 & 32.7 ± 1.6 & 42.3 ± 2.1 & 36.6 ± 2.7 \\
    & CSD~\cite{somepalli2024measuringstyle} & 19.3 ± 1.9 & 37.9 ± 1.4 & 42.4 ± 2.1 & 33.1 ± 2.9 \\ \midrule
\multirow{1}{*}{{\shortstack[c]{Fine-tuned\\classifiers}}}
    & Prototypical Network~\cite{snell2017prototypical} & \textbf{45.1 ± 3.2} & \textbf{77.6 ± 2.1} & \textbf{61.5 ± 2.3} & \textbf{48.3 ± 2.4}  \\ 

    & Vanilla Classifier& 28.5 ± 2.1 & 49.3 ± 1.8 & -- & -- \\ \bottomrule
\end{tabular}
}
\caption{\textbf{Ranked mAP@10 artist retrieval on SDXL 2-artist prompted images}.}
\label{tbl:two_artist_eval}
\end{table*}

\begin{table*}[!t]
\centering
\resizebox{0.9\linewidth}{!}{
\begin{tabular}{llcccccc} \toprule
 & & \multicolumn{5}{c}{Evaluation set} \\ \cmidrule(lr){3-7}

 & & \multicolumn{2}{c}{Seen artists (100-way)} & \multicolumn{2}{c}{Held-out (10-way)} \\ \cmidrule(lr){3-4} \cmidrule(lr){5-6}
 Family & Method & \shortstack[c]{Complex} & \shortstack[c]{Simple} & \shortstack[c]{Complex} & \shortstack[c]{Simple} \\ \midrule
Chance & -- & 3.1 & 3.1 & 45.0 & 45.0 \\ \midrule
\multirow{5}{*}{\shortstack[l]{Retrieval-\\based\\ methods}}
 & AbC - DINO~\cite{wang2023evaluating} & 11.2 ± 1.0 & 20.4 ± 1.0 & 35.1 ± 1.2 & 39.1 ± 1.0 \\
 & AbC - CLIP~\cite{wang2023evaluating} & 10.2 ± 1.0 & 20.2 ± 0.9 & 35.6 ± 1.3 & 43.6 ± 1.3 \\
 & DINOv2~\cite{oquab2023dinov2} & 8.4 ± 0.9 & 13.7 ± 0.9 & 34.3 ± 1.3 & 39.3 ± 1.1 \\
    & CLIP~\cite{radford2021learning} & 11.3 ± 1.1 & 21.3 ± 1.0 & 37.1 ± 1.2 & 46.6 ± 1.3 \\
    & CSD~\cite{somepalli2024measuringstyle} & 13.2 ± 1.1 & 25.0 ± 1.0 & 34.8 ± 1.2 & 41.6 ± 1.1 \\ \midrule
\multirow{1}{*}{{\shortstack[c]{Fine-tuned\\classifiers}}}
    & Prototypical Network~\cite{snell2017prototypical} & \textbf{35.3 ± 2.7} & \textbf{61.9 ± 2.1} & \textbf{62.2 ± 1.3} & \textbf{63.0 ± 1.4}  \\ 
    & Vanilla Classifier& 20.7 ± 1.6 & 34.2 ± 1.4 & -- & -- \\ \bottomrule
\end{tabular}
}
\caption{\textbf{Ranked mAP@10 artist retrieval on SDXL 3-artist prompted images}.}
\label{tbl:three_artist_eval}
\end{table*}

\end{document}